\documentclass{article}

\usepackage{microtype}
\usepackage{graphicx}
\usepackage{subfigure}
\usepackage{booktabs} 

\usepackage[utf8]{inputenc} 
\usepackage[T1]{fontenc}    
\usepackage{hyperref}       
\usepackage{url}            
\usepackage{booktabs}       
\usepackage{amsfonts}       
\usepackage{nicefrac}       
\usepackage{microtype}      
\usepackage{hyperref}
\usepackage{url}
\usepackage{mathtools}
\usepackage{bbm}
\usepackage{enumitem}
\usepackage{amsmath,amssymb,amsthm}
\usepackage{graphicx}
\usepackage{subfigure}
\usepackage[export]{adjustbox}
\usepackage{xcolor}
\usepackage{changes}

\newcommand{\alert}[1]{\textbf{}}

\newcommand{\BEAS}{\begin{eqnarray*}}
\newcommand{\EEAS}{\end{eqnarray*}}
\newcommand{\BEA}{\begin{eqnarray}}
\newcommand{\EEA}{\end{eqnarray}}
\newcommand{\BEQ}{\begin{equation}}
\newcommand{\EEQ}{\end{equation}}
\newcommand{\BIT}{\begin{itemize}}
\newcommand{\EIT}{\end{itemize}}
\newcommand{\BNUM}{\begin{enumerate}}
\newcommand{\ENUM}{\end{enumerate}}
\newcommand{\BEL}[1]{\begin{equation}\label{#1}}
\newcommand{\EEL}{\end{equation}}

\newcommand{\state}{\mathbf{s}}

\newcommand{\action}{\mathbf{a}}

\newcommand{\policy}{\pi}

\newcommand{\reward}{r}
\newcommand{\rewmodel}{\widehat{\reward}_\psi}

\newcommand{\return}{\mathcal{R}}

\newcommand{\BA}{\begin{array}}
\newcommand{\EA}{\end{array}}

\DeclareMathOperator*{\expec}{\mathbb{E}}

\definechangesauthor[color=magenta, name=laura]{LS}
\definechangesauthor[color=purple, name=kimin]{KM}

\newcommand{\methodname}{Unsupervised Pre-training and Preference-Based Learning via Relabeling Experience}

\newcommand{\metabbr}{PEBBLE} 

\usepackage[accepted]{icml2021}

\icmltitlerunning{\methodname}

\begin{document}

\twocolumn[\icmltitle{\metabbr: Feedback-Efficient Interactive Reinforcement Learning \\via Relabeling Experience and Unsupervised Pre-training}

\icmlsetsymbol{equal}{*}

\begin{icmlauthorlist}
\icmlauthor{Kimin Lee}{equal,to}
\icmlauthor{Laura Smith}{equal,to}
\icmlauthor{Pieter Abbeel}{to}
\end{icmlauthorlist}

\icmlaffiliation{to}{University of California, Berkeley}

\icmlcorrespondingauthor{Kimin Lee}{kiminlee@berkeley.edu}
\icmlcorrespondingauthor{Laura Smith}{smithlaura@berkeley.edu}

\icmlkeywords{Machine Learning, ICML}

\vskip 0.3in]



\printAffiliationsAndNotice{\icmlEqualContribution} 

\begin{abstract}
Conveying complex objectives to reinforcement learning (RL) agents can often be difficult, involving meticulous design of reward functions that are sufficiently informative yet easy enough to provide. Human-in-the-loop RL methods allow practitioners to instead interactively teach agents through tailored feedback; however, such approaches have been challenging to scale since human feedback is very expensive. In this work, we aim to make this process more sample- and feedback-efficient. We present an off-policy, interactive RL algorithm that capitalizes on the strengths of both feedback and off-policy learning. Specifically, we learn a reward model by actively querying a teacher's preferences between two clips of behavior and use it to train an agent. To enable off-policy learning, we relabel all the agent's past experience when its reward model changes. We additionally show that pre-training our agents with unsupervised exploration substantially increases the mileage of its queries. We demonstrate that our approach is capable of learning tasks of higher complexity than previously considered by human-in-the-loop methods, including a variety of locomotion and robotic manipulation skills. We also show that our method is able to utilize real-time human feedback to effectively prevent reward exploitation and learn new behaviors that are difficult to specify with standard reward functions. 
\end{abstract}


\section{Introduction}

\begin{figure*} [t] \centering
\includegraphics[width=.99\textwidth]{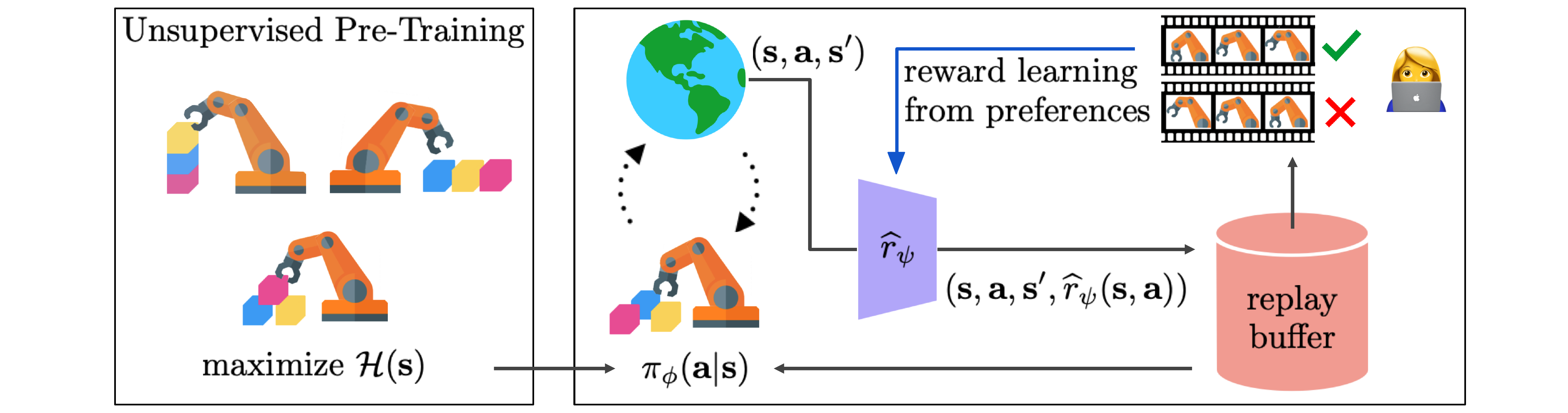}
\caption{Illustration of our method. 
First, the agent engages in unsupervised pre-training during which it is encouraged to visit a diverse set of states so its queries can provide more meaningful signal than on randomly collected experience (left).
Then, a teacher provides preferences between two clips of behavior, and we learn a reward model based on them. The agent is updated to maximize the expected return under the model. We also relabel all its past experiences with this model to maximize their utilization to update the policy (right).}
\label{fig:framework}
\vspace{-0.1in}
\end{figure*}

Deep reinforcement learning (RL) has emerged as a powerful method whereby agents learn complex behaviors on their own through trial and error~\cite{quadrupedal, policy-search, robot-rl, alphago, in-hand-manip, qtopt, alphastar}. Scaling RL to many applications, however, is yet precluded by a number of challenges. One such challenge lies in providing a suitable reward function. 
For example, while it may be desirable to provide sparse rewards out of ease, they are often insufficient to train successful RL agents. Thus, to provide adequately dense signal, real-world problems may require extensive instrumentation, such as accelerometers to detect door opening~\cite{dagps}, thermal cameras to detect pouring~\cite{pouring} or motion capture for object tracking~\cite{em-rl, akkaya2019solving, peng2020learning}. 

Despite these costly measures, it may still be difficult to construct a suitable reward function due to reward exploitation. That is, RL algorithms often discover ways to achieve high returns by unexpected, unintended means. In general, there is nuance in how we might want agents to behave, such as obeying social norms, that are difficult to account for and communicate effectively through an engineered reward function~\citep{amodei2016concrete,implicit-preferences, reward-side-effects}. 
A popular way to avoid reward engineering is through imitation learning, during which a learner distills information about its objectives or tries to directly follow an expert~\citep{schaal1997learning,ng2000irl,abbeel2004irl,argall2009il}. While imitation learning is a powerful tool, suitable demonstrations are often prohibitively expensive to obtain in practice~\cite{collab-manip, online-movement-adaptation, traj-keyframe, vr-teleop}. 

In contrast, humans often learn fairly autonomously, relying on occasional external feedback from a teacher. Part of what makes a teacher effective is their ability to interactively guide students according to their progress, providing corrective or increasingly advanced instructions as needed. Such an interactive learning process is also alluring for artificial agents since the agent's behavior can naturally be tailored to one’s preference (avoiding reward exploitation) without requiring extensive engineering.
This approach is only feasible if the feedback is both practical for a human to provide and sufficiently high-bandwidth. 
As such, human-in-the-loop (HiL) RL~\cite{ tamer, preference_drl,coach} has not yet been widely adopted.

In this work, we aim to substantially reduce the amount of human effort required for HiL learning.
To this end, we present \metabbr: unsupervised \textbf{P}r\textbf{E}-training and preference-\textbf{B}ased learning via rela\textbf{B}e\textbf{L}ing \textbf{E}xperience, a feedback-efficient RL algorithm by which learning is largely autonomous and supplemented by a practical number of binary labels (i.e. preferences) provided by a supervisor.
Our method relies on two main, synergistic ingredients: \emph{unsupervised pre-training} and \emph{off-policy learning} (see Figure~\ref{fig:framework}). 
For generality, we do not assume the agent is privy to rewards from its environment. 
Instead, we first allow the agent to explore using only intrinsic motivation~\citep{oudeyer2007intrinsic,schmidhuber2010formal} to diversify its experience and produce coherent behaviors.
Collecting a breadth of experiences enables the teacher to provide more meaningful feedback, as compared to feedback on data collected in an indeliberate manner. The supervisor then steps in to teach the agent by expressing their preferences between pairs of clips of the agent’s behavior~\citep{preference_drl}. The agent distills this information into a reward model and uses RL to optimize this inferred reward function.

Leveraging unsupervised pre-training increases the efficiency of the teacher’s initial feedback; however, RL requires a large enough number of samples such that supervising the learning process is still quite expensive for humans. It is thus especially critical to enable off-policy algorithms that can reuse data to maximize the agent's, and thereby human's, efficiency. However, on-policy methods have typically been used thus far for HiL RL because of their ability to mitigate the effects of non-stationarity in reward caused by online learning. We show that by simply relabeling all of the agent's past experience every time the reward model is updated, we can make use and reuse of \emph{all} the agent's collected experience
to improve sample and feedback efficiency by a large margin. 
Source code and videos are available at \url{https://sites.google.com/view/icml21pebble}.

We summarize the main contributions of \metabbr:
\begin{itemize} [leftmargin=4mm]
\setlength\itemsep{0.1em}
\item For the first time, we show that \emph{unsupervised pre-training} and \emph{off-policy learning} can significantly improve the sample- and feedback-efficiency of HiL RL.
\item \metabbr~outperforms prior preference-based RL baselines on complex locomotion and robotic manipulation tasks from DeepMind Control Suite (DMControl;~\citealt{dmcontrol_old,dmcontrol_new}) and Meta-world~\citep{yu2020meta}.
\item We demonstrate that \metabbr~can learn behaviors for which a typical reward function is difficult to engineer very efficiently. 
\item We also show that \metabbr~ can avoid reward exploitation, leading to more desirable behaviors compared to an agent trained with respect to an engineered reward function.
\end{itemize}


\section{Related Work}

{\bf Learning from human feedback}. 
Several works have successfully utilized feedback from real humans to train agents where it is assumed that the feedback is available at \textit{all} times~\citep{pilarski2011online,coach,deepcoach}. 
Due to this high feedback frequency, these approaches are difficult to scale to more complex learning problems that require substantial agent experience. 

Better suited to learning in complex domains is to learn a reward model so the agent can learn without a supervisor’s perpetual presence. 
One simple yet effective direction in reward learning is to train a classifier that recognizes task success and use it as basis for a reward function~\citep{pinto2016supersizing, levine2018learning, fu2018variational, xie2018few}. Positive examples may be designated or reinforced through human feedback~\citep{zhang2019solar, singh2019end, smith2020avid}.
Another promising direction has focused on simply training a reward model via regression using unbounded real-valued feedback~\citep{tamer,deeptamer}, but this has been challenging to scale because it is difficult for humans to reliably provide a particular utility value for certain behaviors of the RL agent.

Much easier for humans is to make \emph{relative} judgments, i.e., comparing behaviors as better or worse. Preference-based learning is thus an attractive alternative because the supervision is easy to provide yet information-rich~\cite{akrour2011preference,pilarski2011online,akrour2012april,wilson2012bayesian,sugiyama2012preference, wirth2013preference, wirth2016model, sadigh2017active,  biyik2018batch, leike2018scalable, biyik2020active}. \citet{preference_drl} scaled preference-based learning to utilize modern deep learning techniques---they learn a reward function, modeled with deep neural networks, that is consistent with the observed preferences and use it to optimize an agent using RL. They choose on-policy RL methods~~\citep{trpo,mnih2016a2c} since they are more robust to the non-stationarity in rewards caused by online learning. 
Although they demonstrate that preference-based learning provides a fairly efficient (requiring feedback on less than 1\% of the agent's experience) means of distilling information from feedback, they rely on notoriously sample-inefficient on-policy RL, so a large burden can yet be placed on the human. Subsequent works have aimed to improve the efficiency of this method by introducing additional forms of feedback such as demonstrations~\citep{ibarz2018preference_demo} or non-binary rankings~\citep{cao2020weak}. Our proposed approach similarly focuses on developing a more sample- and feedback-efficient preference-based RL algorithm \emph{without} adding any additional forms of supervision. Instead, we enable off-policy learning as well as utilize \emph{unsupervised} pre-training to substantially improve efficiency.

{\bf Unsupervised pre-training for RL}. 
Unsupervised pre-training has been studied for extracting strong behavioral priors that can be utilized to solve downstream tasks efficiently in the context of RL~\citep{daniel2016hierarchical,florensa2017stochastic,achiam2018valor,EysenbachGIL19diayn,SharmaGLKH20dads}. 
Specifically, 
agents are encouraged to expand the boundary of seen states by maximizing various intrinsic rewards, such as prediction errors~\cite{houthooft2016vime,pathak2017curiosity, burda2018exploration}, count-based state novelty~\cite{bellemare2016unifying, tang2017exploration, ostrovski2017count}, mutual information~\citep{EysenbachGIL19diayn} and state entropy~\cite{hazan2019provably,lee2019efficient,liu2021unsupervised}.
Such unsupervised pre-training methods allow learning diverse behaviors without extrinsic rewards, effectively facilitating accelerated learning of downstream tasks.
In this work, we show that unsupervised pre-training enables a teacher to provide more meaningful signal by showing them a diverse set of behaviors.

\section{Preliminaries} \label{sec:background}

{\bf Reinforcement learning}. 
We consider a standard RL framework where an agent interacts with an environment in discrete time.
Formally, at each timestep $t$, the agent receives a state $\state_t$ from the environment and chooses an action $\action_t$ based on its policy $\policy$.
The environment returns a reward $\reward_t$ 
and the agent transitions to the next state $\state_{t+1}$.
The return $\return_t = \sum_{k=0}^\infty \gamma^k \reward_{t+k}$ is the discounted sum of rewards from timestep $t$ with discount factor $\gamma \in [0,1)$. 
RL then maximizes the expected return from each state $\state_t$.


{\bf Soft Actor-Critic}. 
SAC~\citep{sac} is an off-policy actor-critic method based on the maximum entropy RL framework \citep{ziebart2010modeling}, 
which encourages exploration and greater robustness to noise
by maximizing a weighted objective of the reward and the policy entropy.
To update the parameters, SAC alternates between a soft policy evaluation and a soft policy improvement.
At the soft policy evaluation step,
a soft Q-function, which is modeled as a neural network with parameters $\theta$, is updated by minimizing the following soft Bellman residual:
\begin{align}
  &\mathcal{L}^{\tt SAC}_{\tt critic} 
  =  \mathbb{E}_{\tau_t \sim \mathcal{B}} \Big[\left(Q_\theta(\state_t,\action_t) - \reward_t  -\gamma {\bar V}(\state_{t+1}) \right)^2 \Big],  \label{eq:sac_critic} \\ 
  & \text{with} \quad {\bar V}(\state_t) = \mathbb{E}_{\action_t\sim \pi_\phi} \big[ Q_{\bar \theta} (\state_t,\action_t) - \alpha \log \pi_{\phi} (\action_t|\state_t) \big], \notag
\end{align}
where $\tau_t = (\state_t,\action_t,\state_{t+1},\reward_t)$ is a transition,
$\mathcal{B}$ is a replay buffer,
$\bar \theta$ are the delayed parameters,
and $\alpha$ is a temperature parameter. 
At the soft policy improvement step,
the policy $\pi_\phi$ is updated by minimizing the following objective:
\begin{align}
  \mathcal{L}^{\tt SAC}_{\tt act} = \mathbb{E}_{\state_t\sim \mathcal{B}, \action_{t}\sim \pi_\phi} \Big[ \alpha \log \pi_\phi (\action_t|\state_t) - Q_{ \theta} (\state_{t},\action_{t}) \Big]. \label{eq:sac_actor}
\end{align}
SAC enjoys good sample-efficiency relative to its on-policy counterparts by reusing its past experiences. However, for the same reason, SAC is not robust to a non-stationary reward function. 

{\bf Reward learning from preferences}. 
We follow the basic framework for learning a reward function $\widehat{r}_\psi$ from preferences in which the function is trained to be consistent with observed human feedback~\cite{wilson2012bayesian,preference_drl}. 
In this framework, a segment $\sigma$ is a sequence of observations and actions $\{\state_k,\action_k, ...,\state_{k+H}, \action_{k+H}\}$. We elicit preferences $y$ for segments $\sigma^0$ and $\sigma^1$,
where $y$ is a distribution indicating which segment a human prefers, i.e., $y\in \{(0,1), (1,0), (0.5, 0.5)\}$.
The judgment is recorded in a dataset $\mathcal{D}$ as a triple $(\sigma^0,\sigma^1,y)$.
By following the Bradley-Terry model~\citep{bradley1952rank},
we model a preference predictor using the reward function $\widehat{r}_\psi$ as follows:
\begin{align}
  P_\psi[\sigma^1\succ\sigma^0] = \frac{\exp\sum_{t} \rewmodel(\state_{t}^1, \action_{t}^1)}{\sum_{i\in \{0,1\}} \exp\sum_{t} \rewmodel(\state_{t}^i, \action_{t}^i)},\label{eq:pref_model}
\end{align}
where $\sigma^i\succ\sigma^j$ denotes the event that segment $i$ is preferable to segment $j$.
Intuitively, this can be interpreted as assuming the probability of preferring a segment depends exponentially on the sum over the segment of an underlying reward function.
While $\rewmodel$ is not a binary classifier, learning $\rewmodel$ amounts to binary classification with labels $y$ provided by a supervisor. Concretely, the reward function, modeled as a neural network with parameters $\psi$, is updated by minimizing the following loss:
\begin{align}
  \mathcal{L}^{\tt Reward} =  -\expec_{(\sigma^0,\sigma^1,y)\sim \mathcal{D}} \Big[ & y(0)\log P_\psi[\sigma^0\succ\sigma^1] \notag \\
  &+ y(1) \log P_\psi[\sigma^1\succ\sigma^0]\Big]. \label{eq:reward-bce}
\end{align}

\section{\metabbr}
In this section, we present \metabbr: unsupervised \textbf{P}r\textbf{E}-training and preference-\textbf{B}ased learning via rela\textbf{B}e\textbf{L}ing \textbf{E}xperience, 
an off-policy actor-critic algorithm for HiL RL. 
Formally, we consider a policy $\pi_\phi$, Q-function $Q_\theta$ and reward function $\rewmodel$, which are updated by the following processes (see Algorithm~\ref{alg:pebble} for the full procedure):
\begin{itemize} [leftmargin=8mm]
\setlength\itemsep{0.1em}
\item {\em Step 0 (unsupervised pre-training)}: We pre-train the policy $\pi_\phi$ only using intrinsic motivation to explore and collect diverse experiences (see Section~\ref{sec:unsuper_exploration}).
\item {\em Step 1 (reward learning)}: We learn a reward function $\rewmodel$ that can lead to the desired behavior by getting feedback from a teacher (see Section~\ref{sec:main_reward_learning}).
\item {\em Step 2 (agent learning)}: We update the policy $\pi_\phi$ and Q-function $Q_\theta$ using an off-policy RL algorithm with 
relabeling to mitigate the effects of a non-stationary reward function $\rewmodel$ (see Section~\ref{sec:agent_learning}).
\item Repeat {\em Step 1} and {\em Step 2}. 
\end{itemize}
\subsection{Accelerating Learning via Unsupervised Pre-training} \label{sec:unsuper_exploration}

In our setting, we assume the agent is given feedback in the form of preferences between segments. In the beginning of training, though, a naive agent executes a random policy, which does not provide good state coverage nor coherent behaviors. The agent's queries are thus quite limited and likely difficult for human teachers to judge. As a result, it requires many samples (and thus queries) for these methods to show initial progress. Recent work has addressed this issue by means of providing demonstrations; however, this is not ideal since these are notoriously hard to procure~\cite{ibarz2018preference_demo}.
Instead, our insight is to produce informative queries at the start of training by utilizing unsupervised pretraining to collect diverse samples solely through intrinsic motivation~\citep{oudeyer2007intrinsic,schmidhuber2010formal}.

\begin{algorithm}[t]
\caption{\texttt{EXPLORE}: Unsupervised exploration 
} \label{alg:unsuprl}
\begin{algorithmic}[1]
\footnotesize
\STATE Initialize parameters of $Q_\theta$ and $\pi_\phi$ and  a replay buffer $\mathcal{B} \leftarrow \emptyset$
\FOR{each iteration}
\FOR{each timestep $t$}
\STATE  Collect $\state_{t+1}$ by taking $\action_t \sim \policy_\phi \left(\action_t | \state_t\right)$
\STATE Compute intrinsic reward $r_t^{\text{\tt int}}\leftarrow r^{\text{\tt int}}(\state_t)$ as in \eqref{eq:intrinsic_reward}
\STATE Store transitions $\mathcal{B} \leftarrow \mathcal{B}\cup \{(\state_t,\action_t,\state_{t+1},r_t^{\text{\tt int}})\}$
\ENDFOR
\FOR{each gradient step}
\STATE Sample minibatch $\{\left(\state_j,\action_j,\state_{j+1},r_j^{\text{\tt int}}\right)\}_{j=1}^B\sim\mathcal{B}$
\STATE Optimize $\mathcal{L}^{\tt SAC}_{\tt critic}$ in \eqref{eq:sac_critic} and $\mathcal{L}^{\tt SAC}_{\tt act}$ in \eqref{eq:sac_actor} with respect to $\theta$ and $\phi$
\ENDFOR
\ENDFOR
\STATE $\textbf{return} \; \mathcal{B}, \policy_\phi$
\end{algorithmic}
\end{algorithm}

Specifically, we encourage our agent to visit a wider range of states by using the state entropy $\mathcal{H}(\state) = -\mathbb{E}_{\state \sim p(\state)} \left[ \log p(\state) \right]$ as an intrinsic reward~\cite{hazan2019provably,lee2019efficient,liu2021unsupervised,seo2021state}.
By updating the agent to maximize the sum of expected intrinsic rewards,
it can efficiently explore an environment and learn how to generate diverse behaviors.
However, this intrinsic reward is intractable to compute in most settings.
To handle this issue,
we employ the simplified version of particle-based entropy estimator~\cite{beirlant1997nonparametric, singh2003nearest} (see the supplementary material for more details):
\begin{align*}
    \widehat{\mathcal{H}}(\state) \propto \sum_i \log (|| \state_{i} - \state_{i}^{k}||),
\end{align*}
where $\widehat{\mathcal{H}}$ denotes the particle-based entropy estimator and $\state_{i}^{k}$ is the $k$-th nearest neighbor ($k$-NN) of $\state_{i}$.
This implies that maximizing the distance between a state and its nearest neighbor increases the overall state entropy. 
Inspired by this, 
we define the intrinsic reward of the current state $\state_t$ as the distance between $\state_{t}$ and its $k$-th nearest neighbor by following the idea of \citet{liu2021unsupervised} that treats each transition as a particle:
\begin{align}
    r^{\text{\tt int}}(\state_t) = \log (|| \state_{t} - \state_{t}^{k}||).
    \label{eq:intrinsic_reward}
\end{align}
In our experiments, we compute $k$-NN distances between a sample and all samples in the replay buffer and normalize the intrinsic reward by dividing it by a running estimate of the standard deviation.
The full procedure of unsupervised pre-training is summarized in Algorithm~\ref{alg:unsuprl}.

\subsection{Selecting Informative Queries} \label{sec:main_reward_learning}

As previously mentioned, we learn our reward function by modeling the probability that a teacher prefers one sampled segment over another as proportional to the exponentiated sum of rewards over the segment (see Section~\ref{sec:background}). Ideally, one should solicit preferences so as to maximize \textit{expected value of information} (EVOI;~\citealt{savage1972foundations}): the improvement of an agent caused by optimizing with respect to the resulting reward model~\cite{viappiani2012monte, akrour2012april}. Computing the EVOI is intractable since it involves taking an expectation over all possible trajectories induced by the updated policy. To handle this issue, several approximations have been explored by prior works to sample queries that are likely to change the reward model~\cite{daniel2015active,preference_drl,ibarz2018preference_demo}.
In this work, we consider the sampling schemes employed by~\citet{preference_drl}: (1) uniform sampling and (2) ensemble-based sampling, which selects pairs of segments with high variance across ensemble reward models. We explore an additional third method, entropy-based sampling, which seeks to disambiguate pairs of segments nearest the decision boundary. That is, we sample a large batch of segment pairs and select pairs that maximize $\mathcal{H}(P_\psi)$. We evaluate the effects of these sampling methods in Section~\ref{sec:experiments}.


\begin{algorithm}[t!]
\caption{\metabbr} \label{alg:pebble}
\begin{algorithmic}[1]
\footnotesize
\REQUIRE frequency of teacher feedback $K$
\REQUIRE number of queries $M$ per feedback session
\STATE Initialize parameters of $Q_\theta$ and $\rewmodel$
\STATE Initialize a dataset of preferences $\mathcal{D} \leftarrow \emptyset$
\STATE {{\textsc{// Exploration phase}}}
\STATE $\mathcal{B}, \policy_\phi \leftarrow\texttt{EXPLORE}()$ in Algorithm~\ref{alg:unsuprl}
\STATE {{\textsc{// Policy learning}}}
\FOR{each iteration}
\STATE {{\textsc{// Reward learning}}}
\IF{iteration \% $K == 0$}
\FOR{$m$ in $1\ldots M$}
\STATE $(\sigma^0, \sigma^1)\sim\texttt{SAMPLE()}$ (see Section~\ref{sec:main_reward_learning})
\STATE Query instructor for $y$
\STATE Store preference $\mathcal{D} \leftarrow \mathcal{D}\cup \{(\sigma^0, \sigma^1,y)\}$
\ENDFOR
\FOR{each gradient step}
\STATE Sample minibatch $\{(\sigma^0, \sigma^1,y)_j\}_{j=1}^D\sim\mathcal{D}$
\STATE Optimize $\mathcal{L}^{\tt Reward}$ in \eqref{eq:reward-bce} with respect to $\psi$ 
\ENDFOR
\STATE Relabel entire replay buffer $\mathcal{B}$ using $\rewmodel$ 
\ENDIF
\FOR{each timestep $t$}
\STATE  Collect $\state_{t+1}$ by taking $\action_t \sim \policy_\phi(\action_t | \state_t)$
\STATE Store transitions $\mathcal{B} \leftarrow \mathcal{B}\cup \{(\state_t,\action_t,\state_{t+1},\rewmodel(\state_t))\}$
\ENDFOR
\FOR{each gradient step}
\STATE Sample random minibatch $\{\left(\tau_j\right)\}_{j=1}^B\sim\mathcal{B}$
\STATE Optimize $\mathcal{L}^{\tt SAC}_{\tt critic}$ in \eqref{eq:sac_critic} and $\mathcal{L}^{\tt SAC}_{\tt act}$ in \eqref{eq:sac_actor} with respect to $\theta$ and $\phi$, respectively
\ENDFOR
\ENDFOR
\end{algorithmic}
\end{algorithm}
\vspace{-0.1in}

\subsection{Using Off-policy RL with Non-Stationary Reward} \label{sec:agent_learning}


Once we learn a reward function ${\widehat r}_\psi$,
we can update the policy $\pi_\phi$ and Q-function $Q_\theta$ using any RL algorithm. A caveat is that the reward function ${\widehat r}_\psi$ may be non-stationary because we update it during training.
\citet{preference_drl} used on-policy RL algorithms, TRPO~\citep{trpo} and A2C~\citep{mnih2016a2c}, to address this issue. However, their poor sample-efficiency leads to poor feedback-efficiency of the overall HiL method.

In this work, we use an off-policy RL algorithm, which provides for sample-efficient learning by reusing past experiences that are stored in the replay buffer.
However,
the learning process of off-policy RL algorithms can be unstable 
because previous experiences in the replay buffer are labeled with previous learned rewards.
To handle this issue,
we relabel all of the agent’s past experience every time we update the reward model.
We find that this simple technique stabilizes the learning process and provides large gains in performance (see Figure~\ref{fig:ablation_unsuper} for supporting results).
The full procedure of\metabbr~ is summarized in Algorithm~\ref{alg:pebble}.


\begin{figure} [t] \centering
\includegraphics[width=0.45\textwidth]{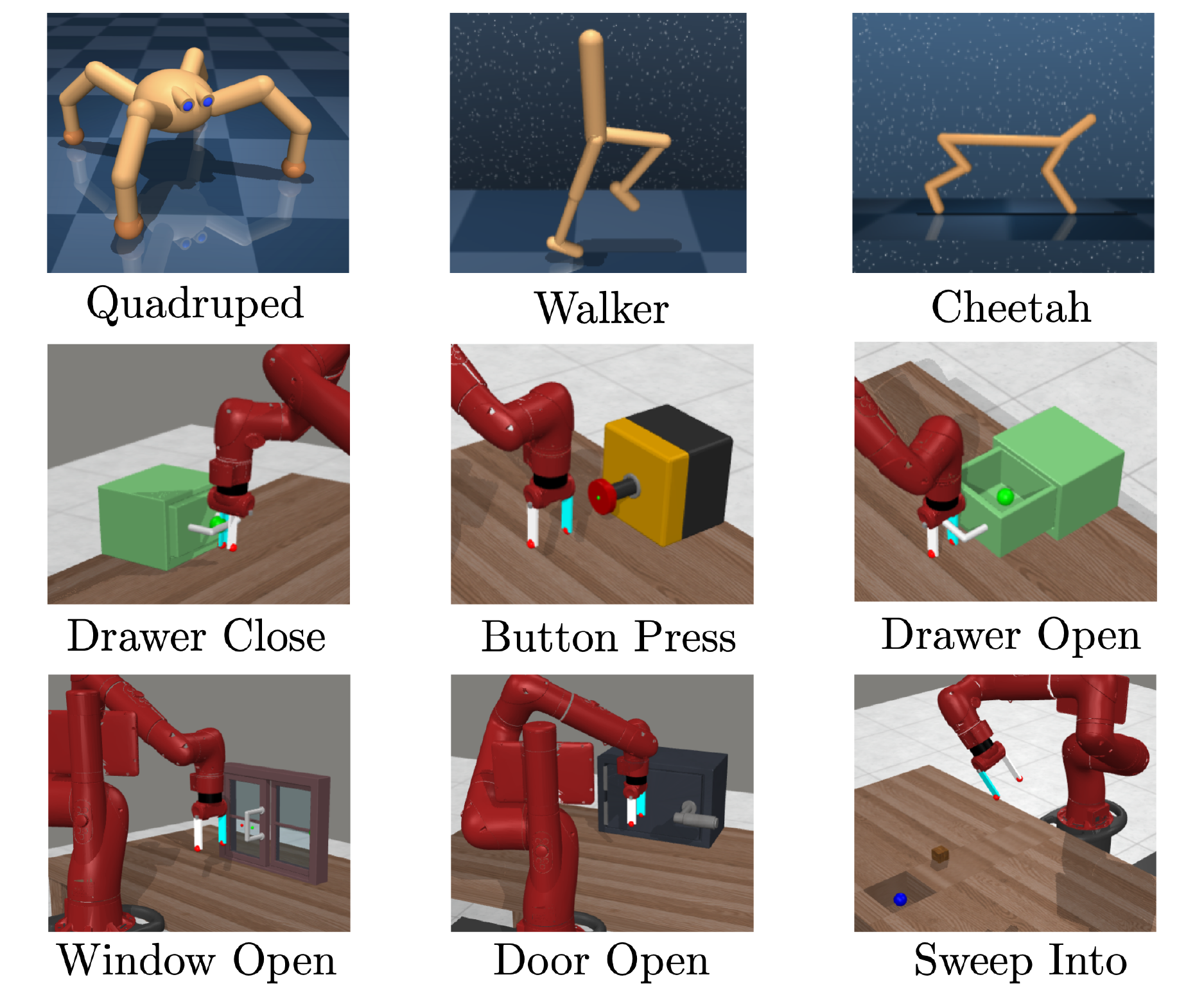}
\caption{Examples from the environments we test on. We consider learning a variety of complex locomotion and manipulation skills through interacting with a scripted or human trainer.}
\label{fig:env_examples}
\end{figure}



\begin{figure*} [t] \centering
\includegraphics[width=.95\textwidth]{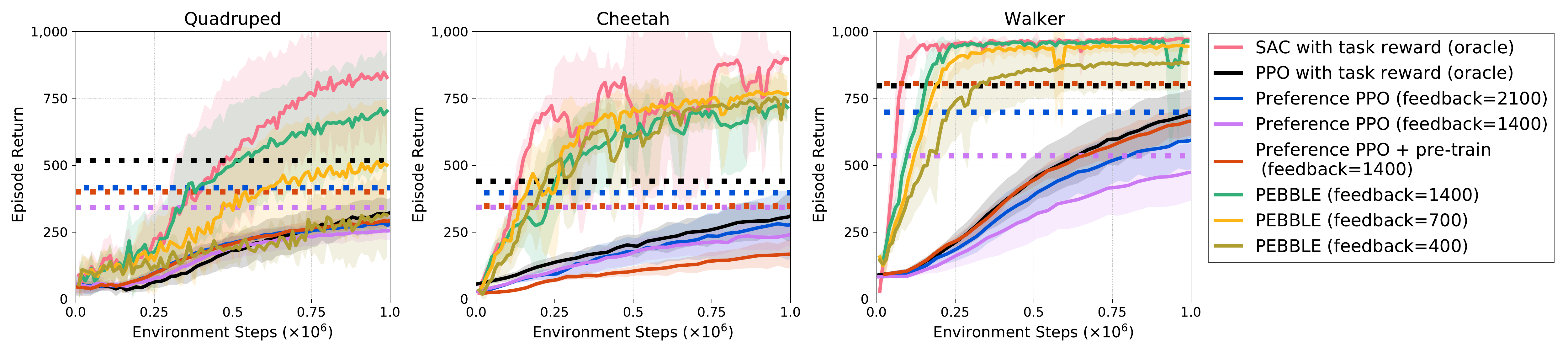}
\caption{Learning curves on locomotion tasks as measured on the ground truth reward. The solid line and shaded regions represent the mean and standard deviation, respectively, across ten runs. Asymptotic performance of PPO and Preference PPO is indicated by dotted lines of the corresponding color.}
\label{fig:main_locomotion}
\end{figure*}

\section{Experiments}\label{sec:experiments}
We design our experiments to investigate the following:
\vspace{-0.1in}
\begin{enumerate} [leftmargin=8mm]
\setlength\itemsep{0.1em}
\item How does \metabbr~compare to existing methods in terms of sample and feedback efficiency?
\item What is the contribution of each of the proposed techniques in \metabbr?
\item Can \metabbr~learn novel behaviors for which a typical reward function is difficult to engineer?
\item Can \metabbr~mitigate the effects of reward exploitation?
\end{enumerate}

\subsection{Setups}

We evaluate \metabbr\, on several continuous control tasks involving locomotion and robotic manipulation
from DeepMind Control Suite (DMControl;~\citealt{dmcontrol_old,dmcontrol_new}) and Meta-world~\citep{yu2020meta}.
In order to verify the efficacy of our method, 
we first focus on having an agent solve a range of 
tasks without being able to directly observe the ground truth reward function.
Instead, similar to \citet{preference_drl} and \citet{ibarz2018preference_demo},
the agent learns to perform a task only by getting feedback from a scripted teacher that provides preferences between trajectory segments according to the true, underlying task reward.
Because this scripted teacher's preferences are immediately generated by a ground truth reward,
we are able
to evaluate the agent quantitatively by measuring the true average return and do more rapid experiments.
For all experiments, we report the mean and standard deviation across ten runs.

We also run experiments with actual human trainers (the authors) to show the benefits of human-in-the-loop RL.
First, we show that human trainers can teach novel behaviors (e.g., waving a leg), which are not defined in original benchmarks.
Second, we show that agents trained with the hand-engineered rewards from benchmarks can perform the task in an undesirable way (i.e., the agent exploits a misspecified reward function), while agents trained using human feedback can perform the same task in the desired way. 
For all experiments, each trajectory segment is presented to the human as a 1 second video clip,
and a maximum of one hour of human time is required.


\begin{figure*}
\centering
\begin{tabular}{c|c|c}
\includegraphics[width=0.3\linewidth]{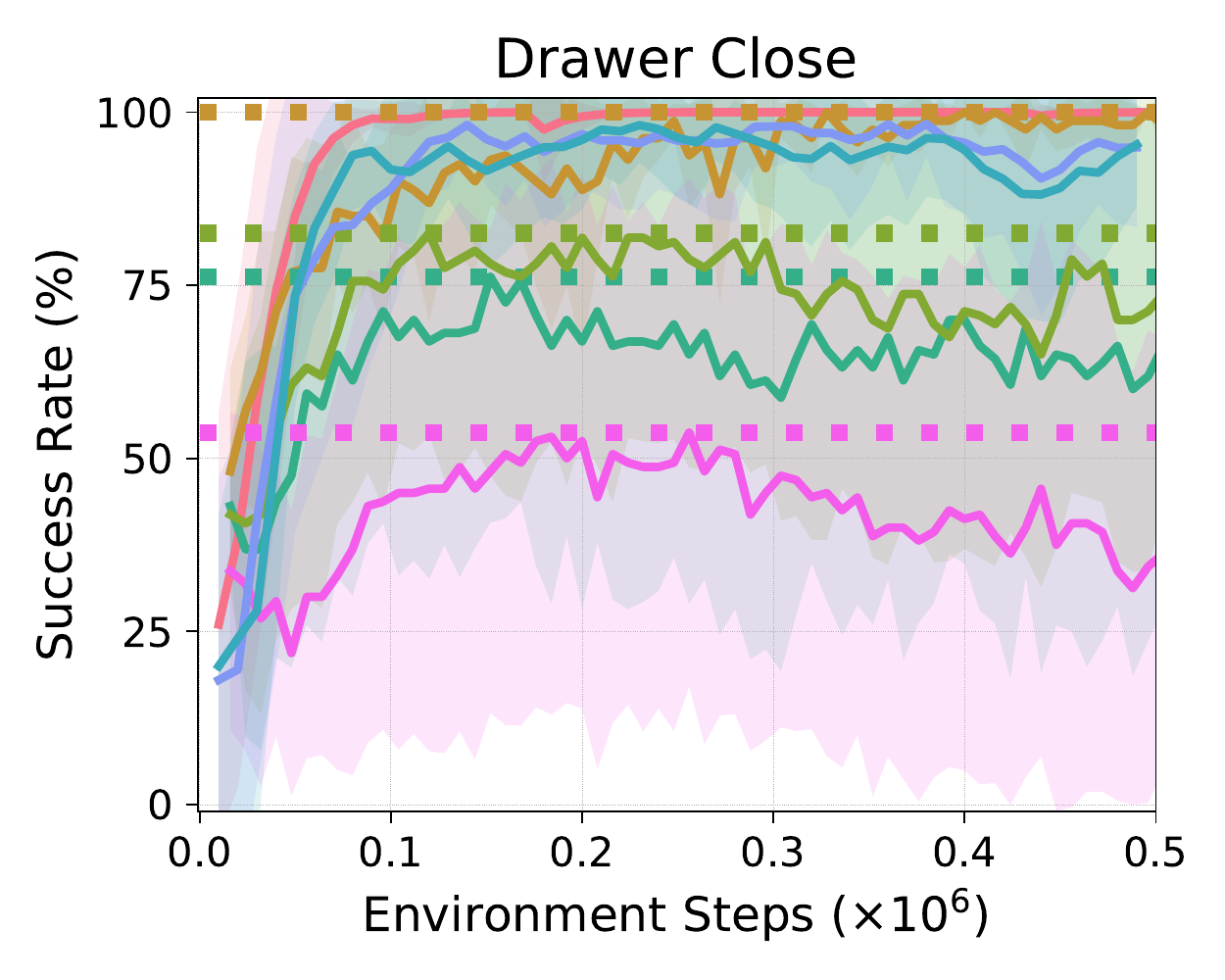}
& \includegraphics[width=0.30\linewidth]{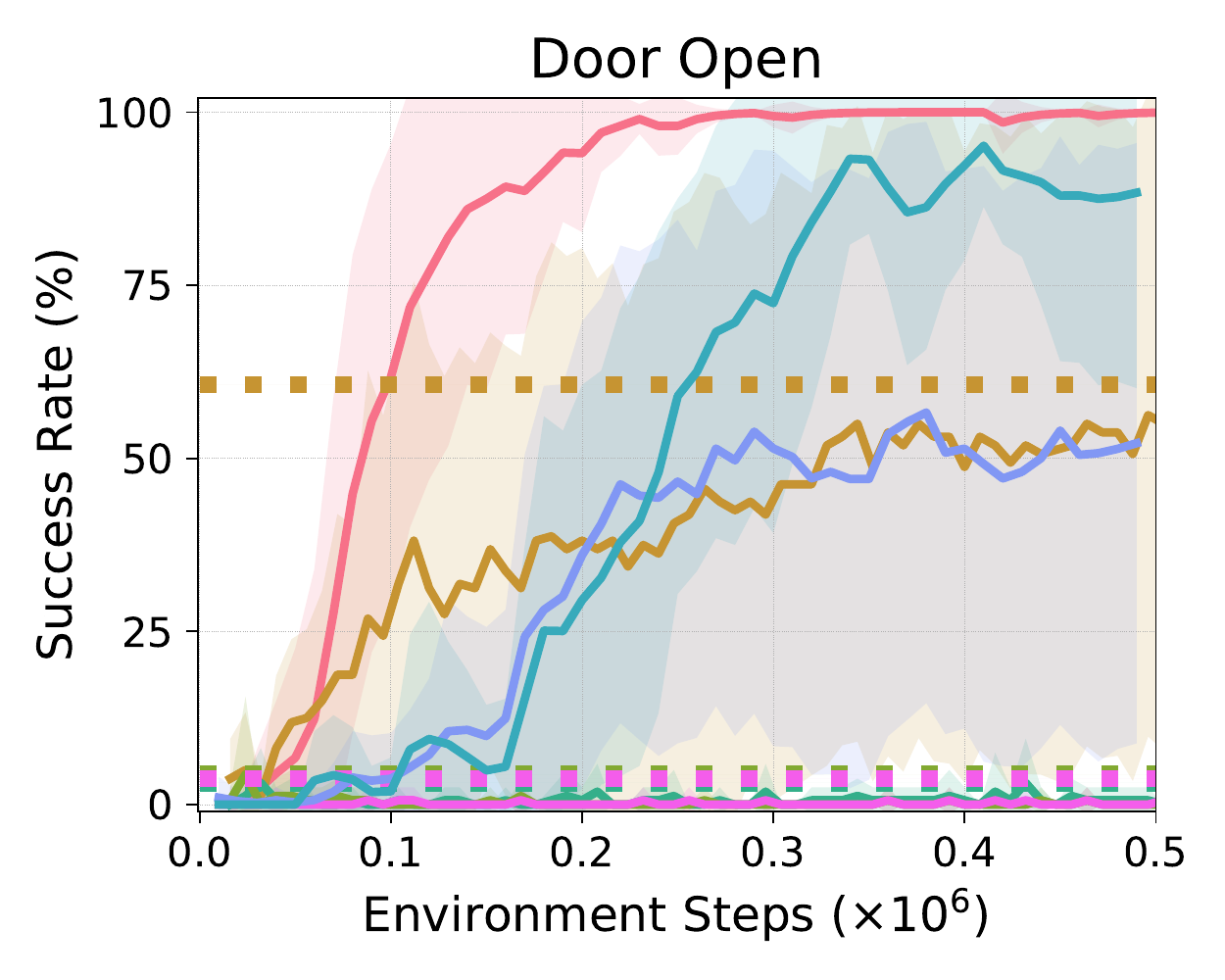} 
& \includegraphics[width=0.30\linewidth]{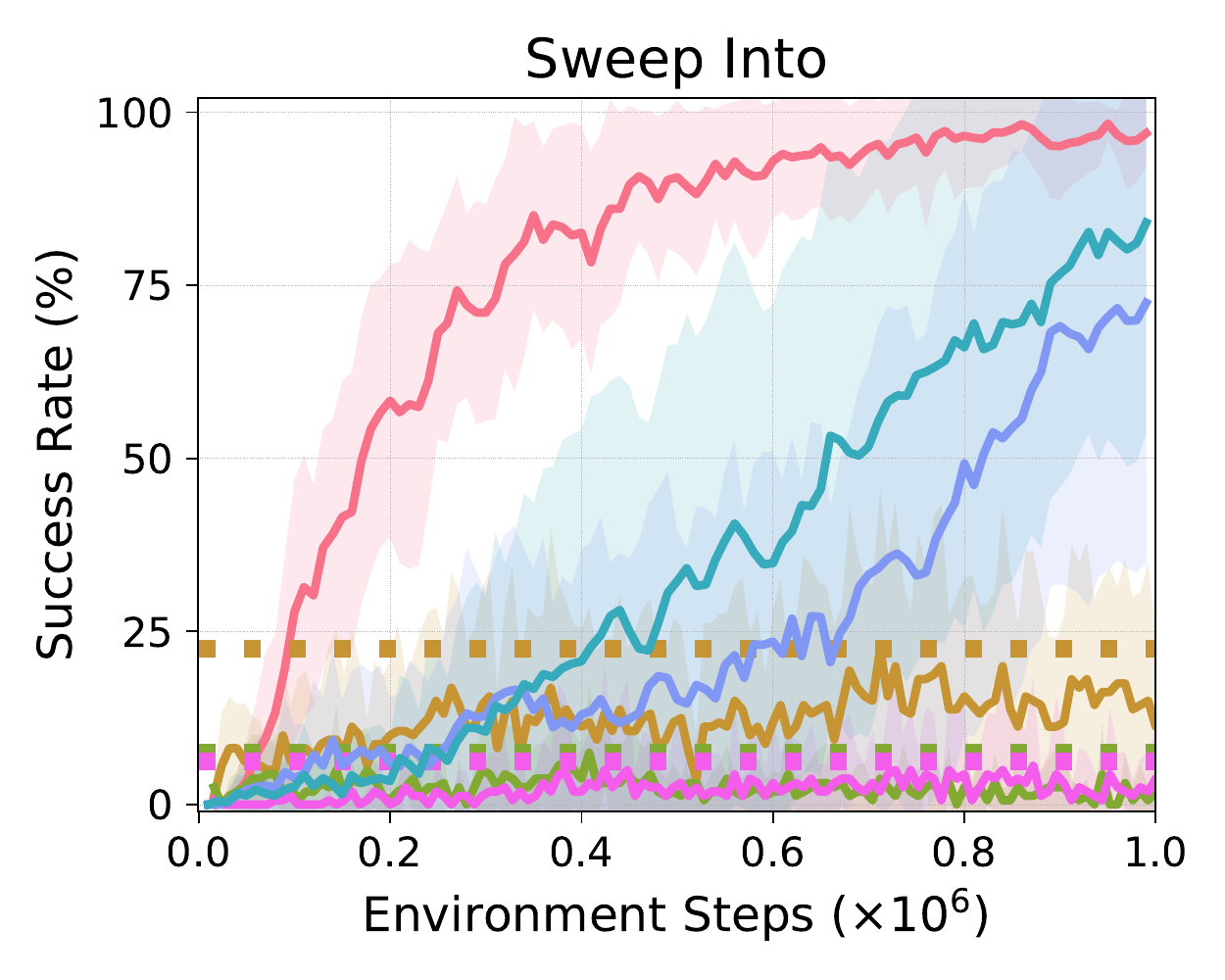} \vspace{-.3cm}\\
\includegraphics[width=0.30\linewidth]{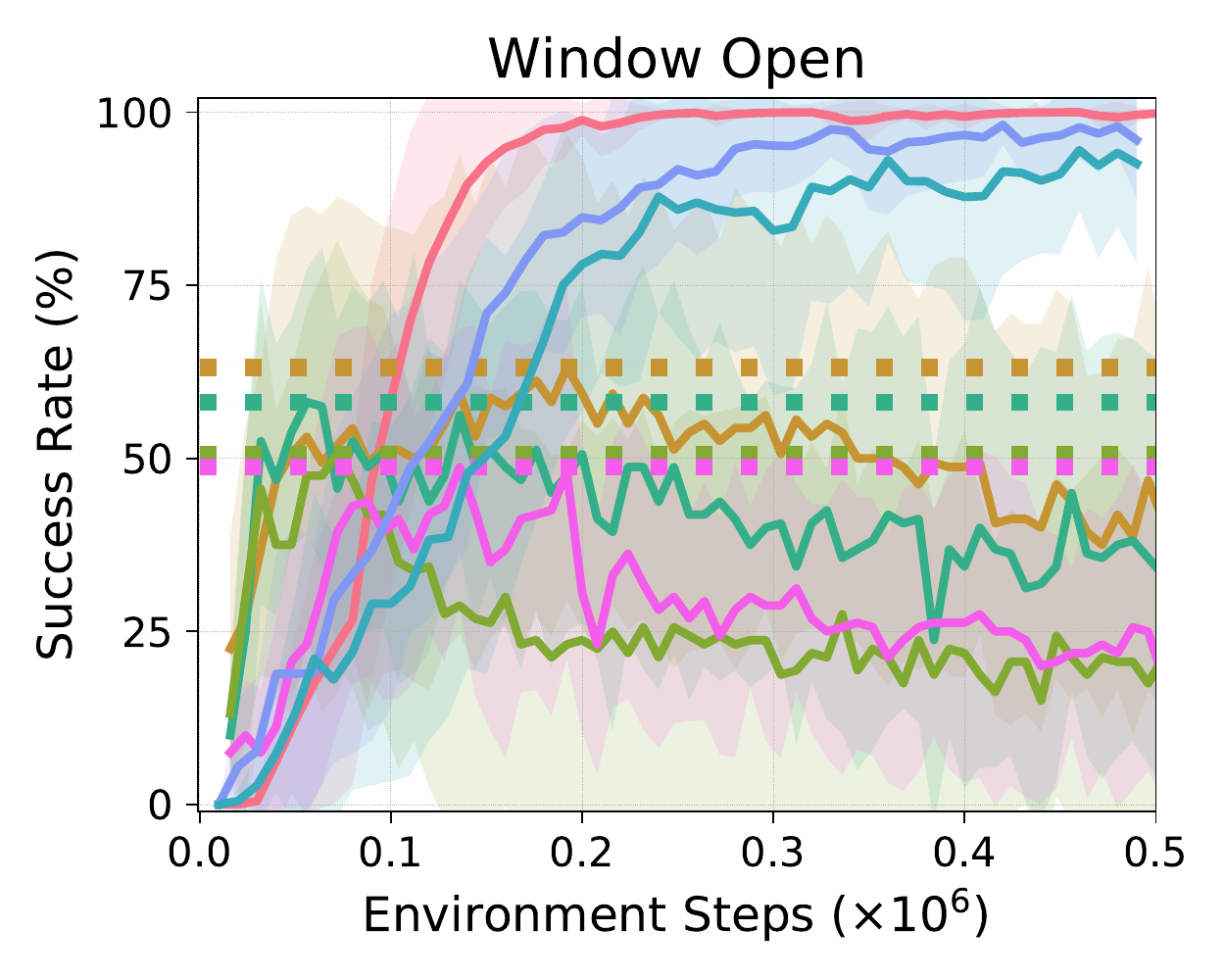} 
& \includegraphics[width=0.30\linewidth]{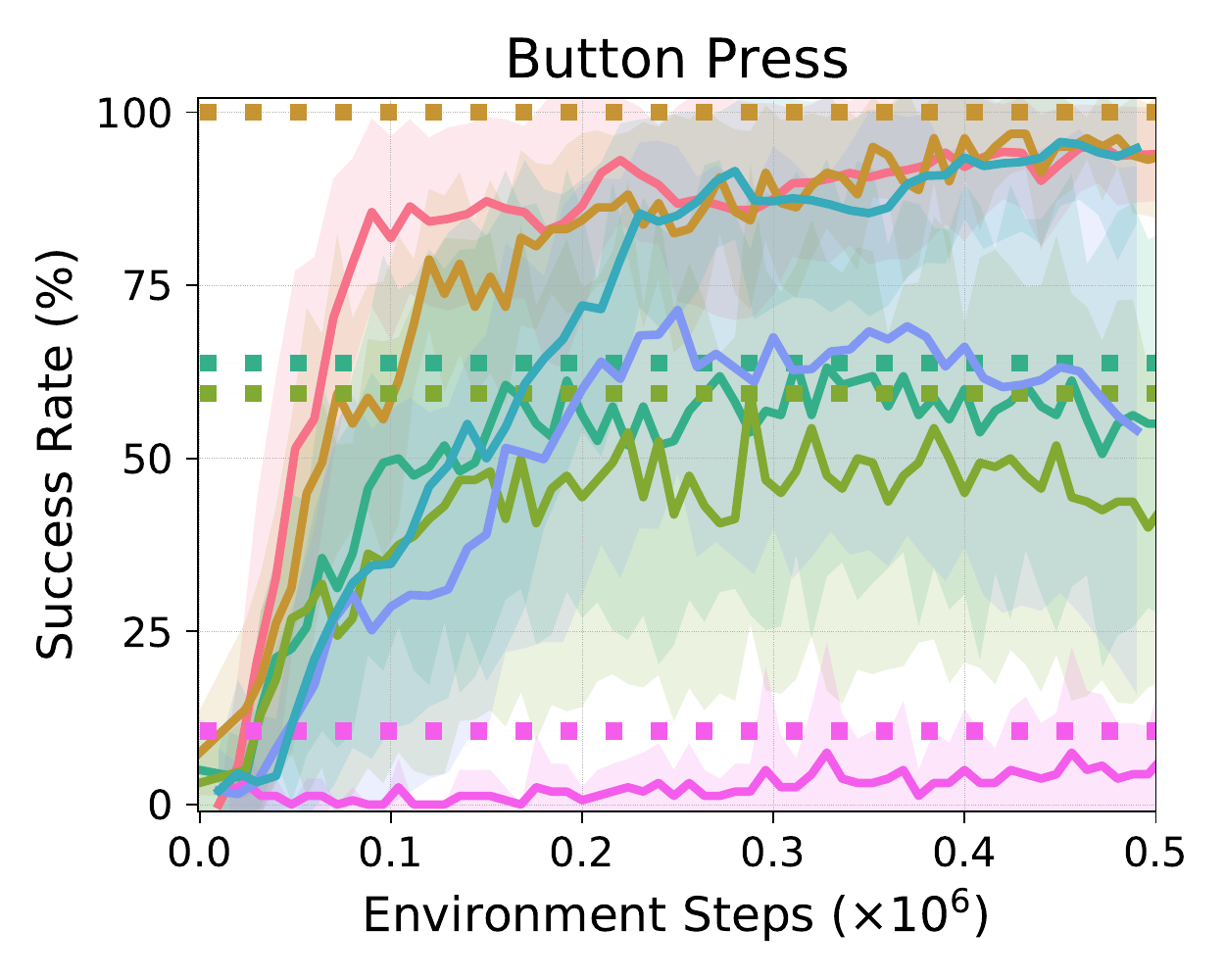} 
& \includegraphics[width=0.3\linewidth]{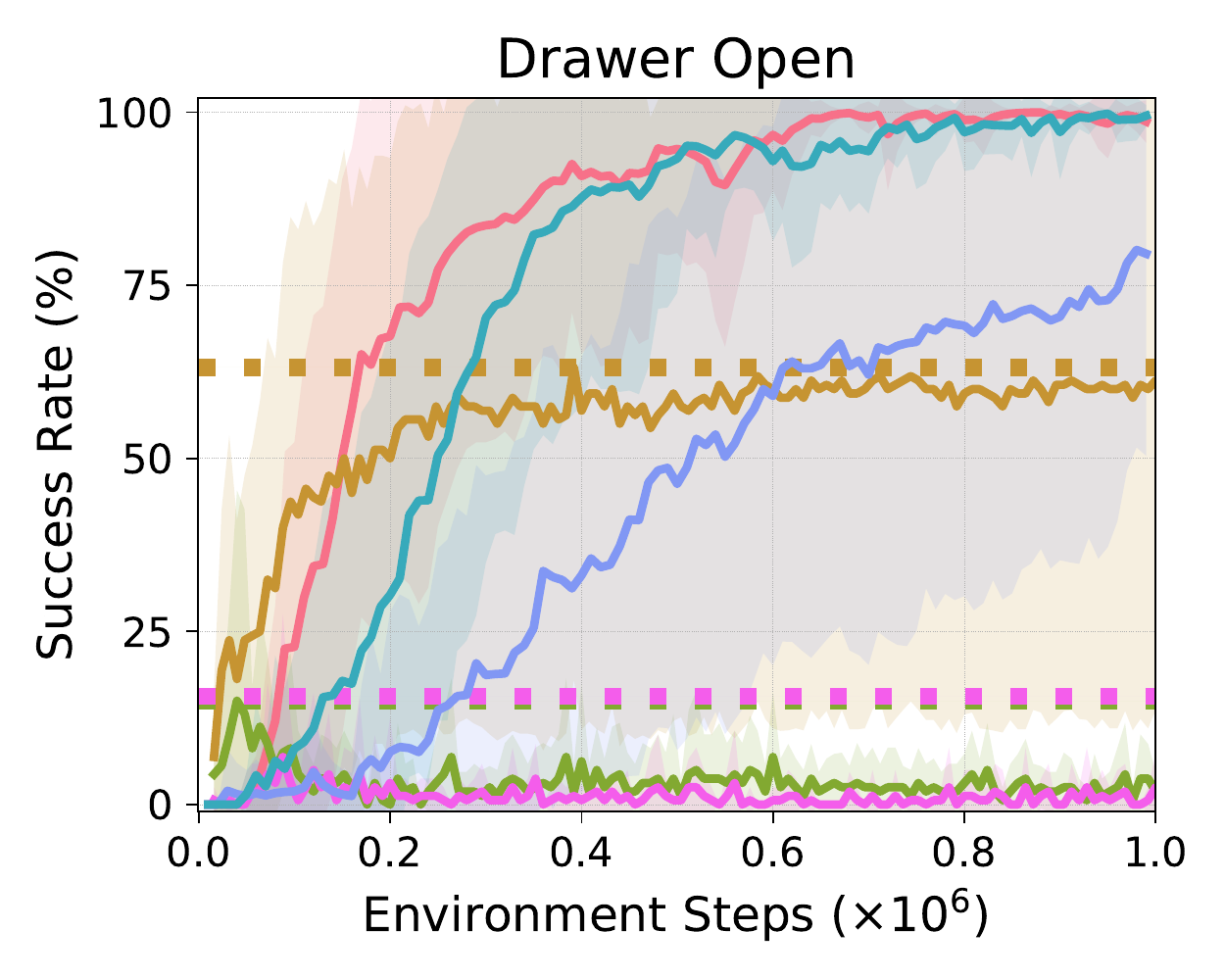}\vspace{-.28cm}\\
\includegraphics[width=0.3\linewidth]{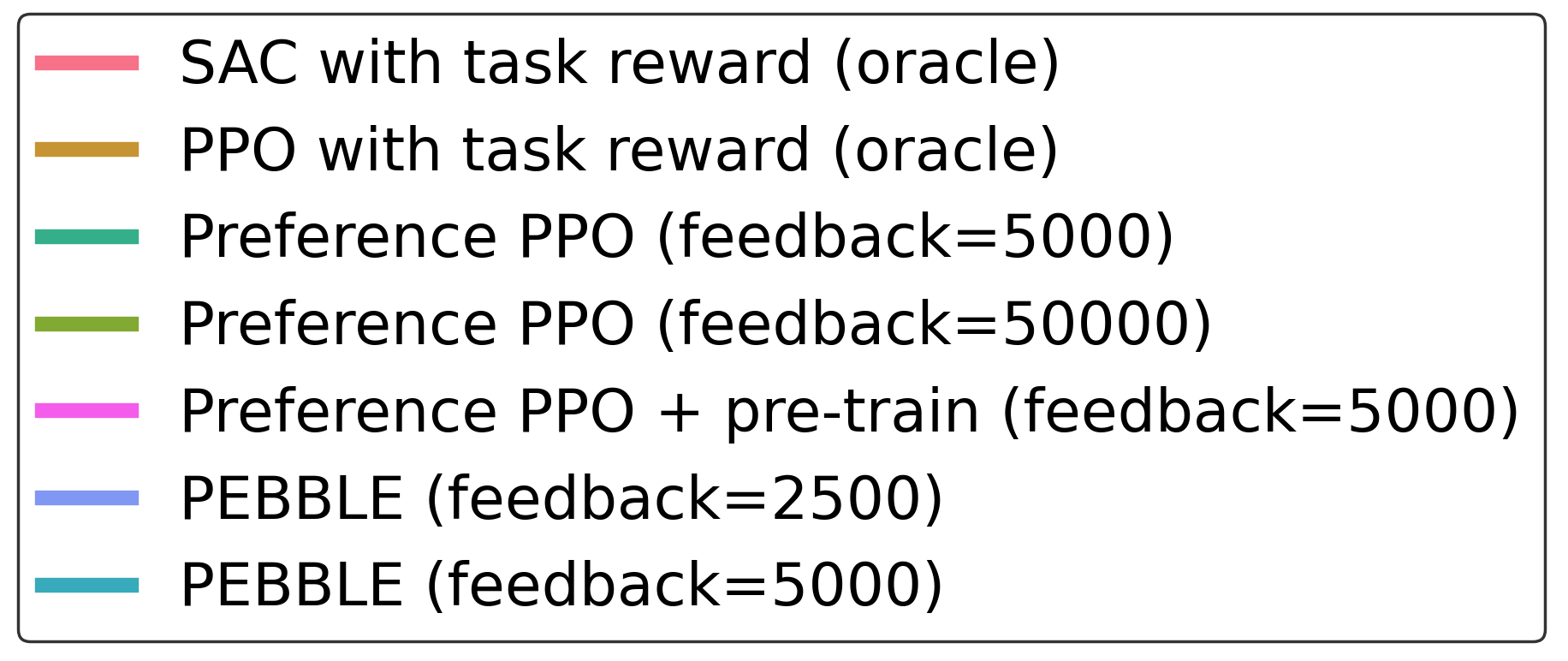}
& \includegraphics[width=0.3\linewidth]{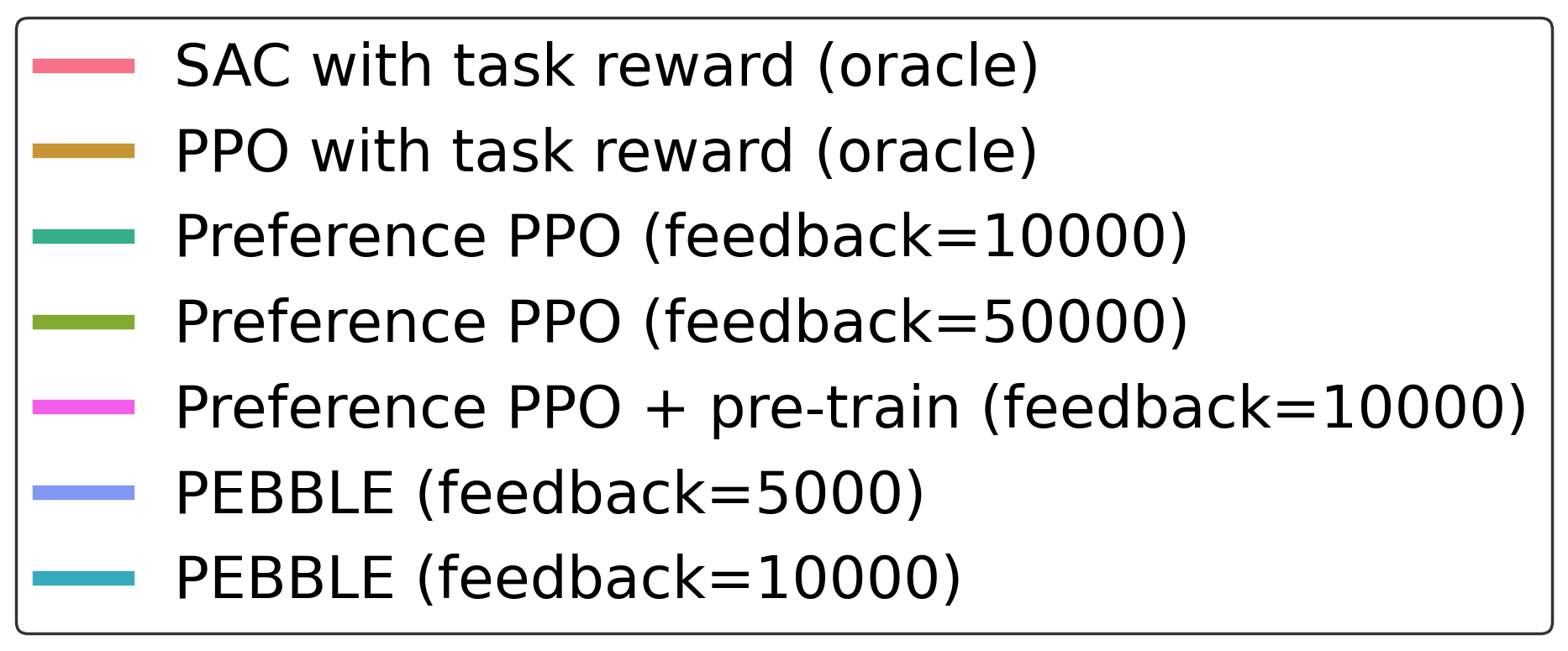} 
& \includegraphics[width=0.3\linewidth]{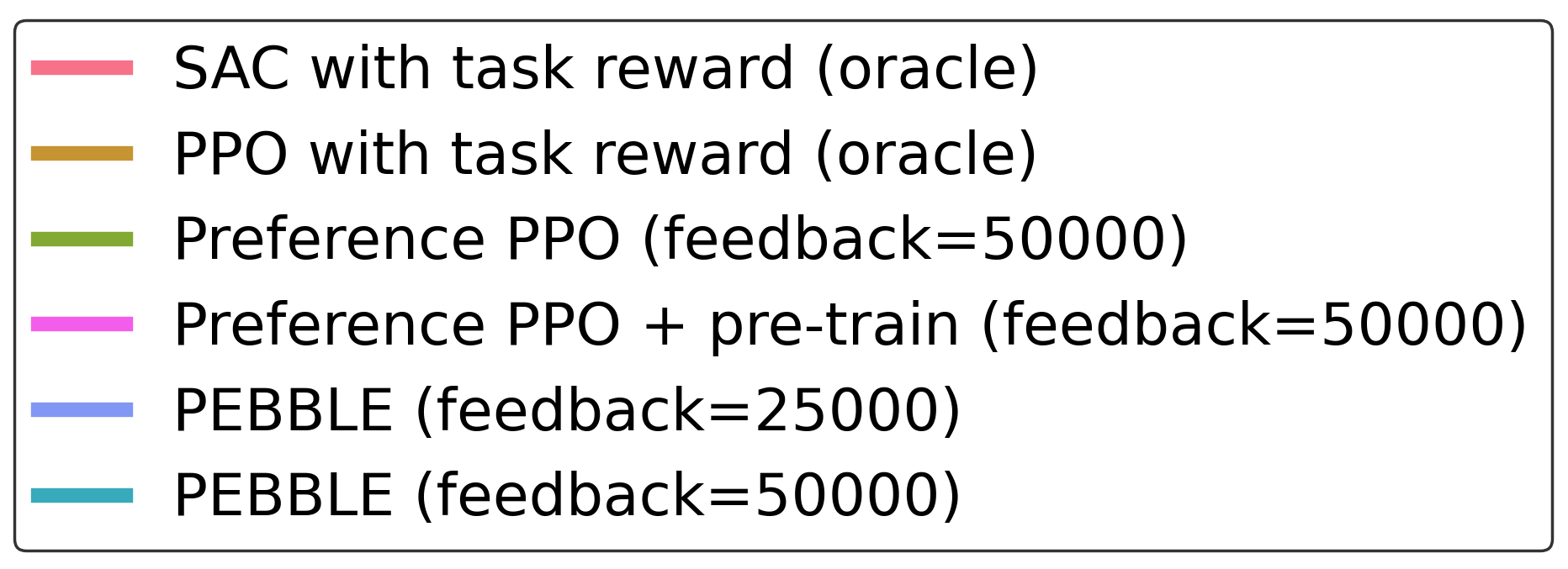}
\vspace{-.25cm}
\end{tabular}
\caption{\footnotesize Learning curves on robotic manipulation tasks as measured on the success rate. The solid line and shaded regions represent the mean and standard deviation, respectively, across ten runs. Asymptotic performance of PPO and Preference PPO is indicated by dotted lines of the corresponding color.}
\label{fig:main_manipulation}
\end{figure*}

\begin{figure*} [t!] \centering
\subfigure[Effects of relabeling and pre-training]
{
\includegraphics[width=0.31\textwidth]{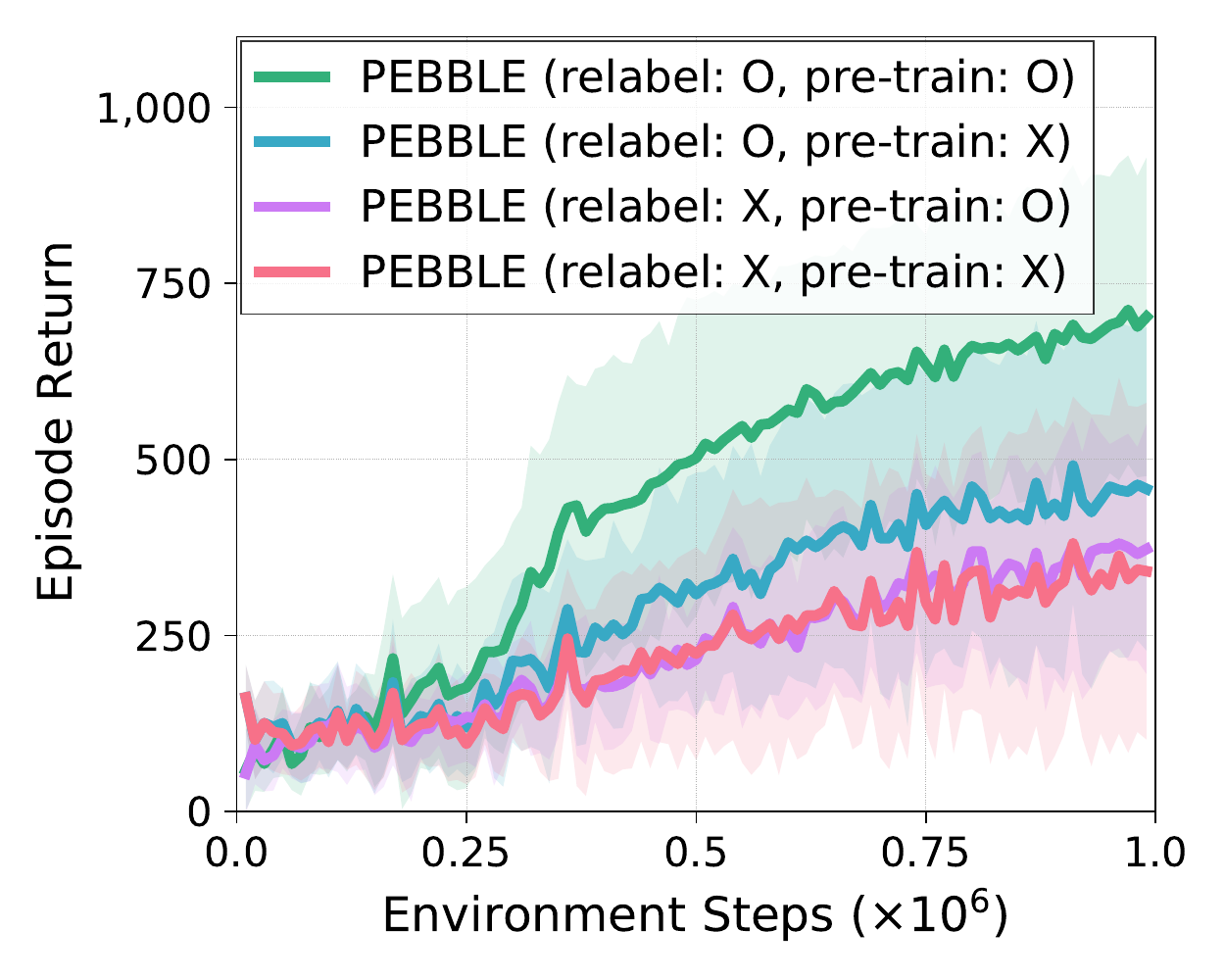}
\label{fig:ablation_unsuper}} 
\subfigure[Sampling schemes]
{
\includegraphics[width=0.31\textwidth]{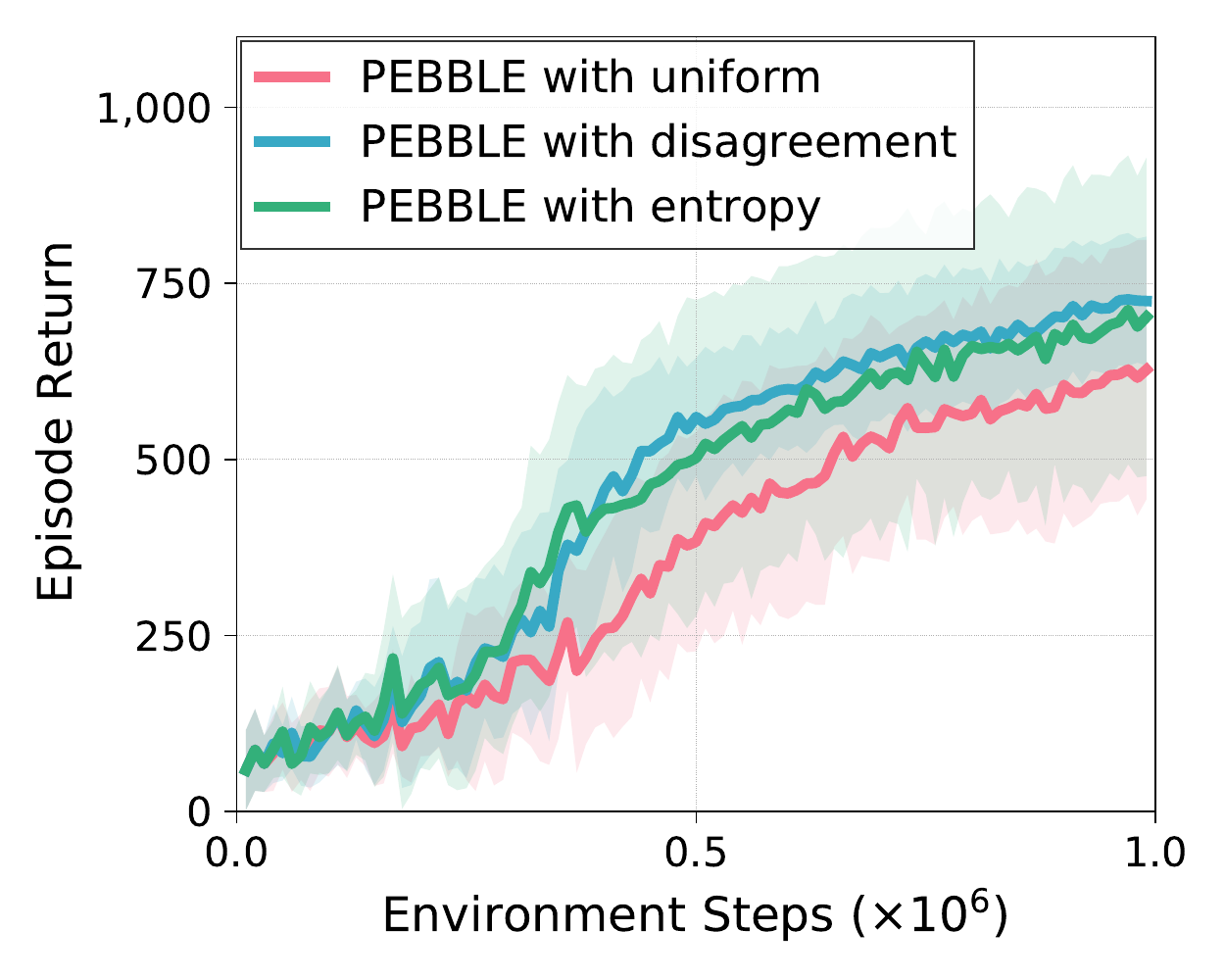}
\label{fig:ablation_sampling}}
\subfigure[Length of the segment]
{
\includegraphics[width=0.31\textwidth]{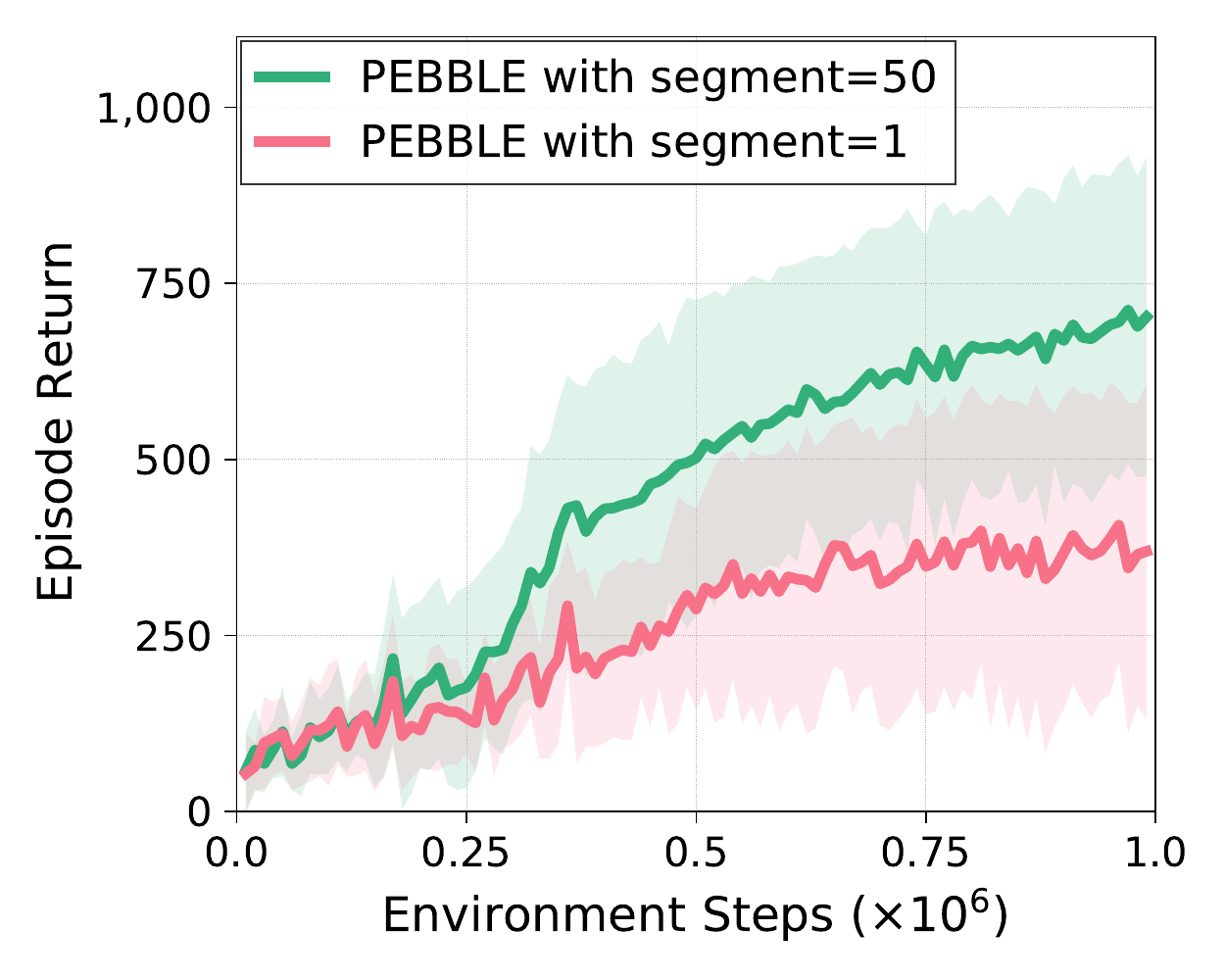}
\label{fig:ablation_seg}}
\caption{Ablation study on Quadruped-walk. 
(a) Contribution of each technique in \metabbr, i.e., relabeling the replay buffer (relabel) and unsupervised pre-training (pre-train). 
(b) Effects of sampling schemes to select queries. 
(c) \metabbr~with varying the length of the segment.
The results show the mean and standard deviation averaged over ten runs.}
\label{fig:ablation}
\end{figure*}

\begin{figure*} [t] \centering
\includegraphics[width=.95\textwidth]{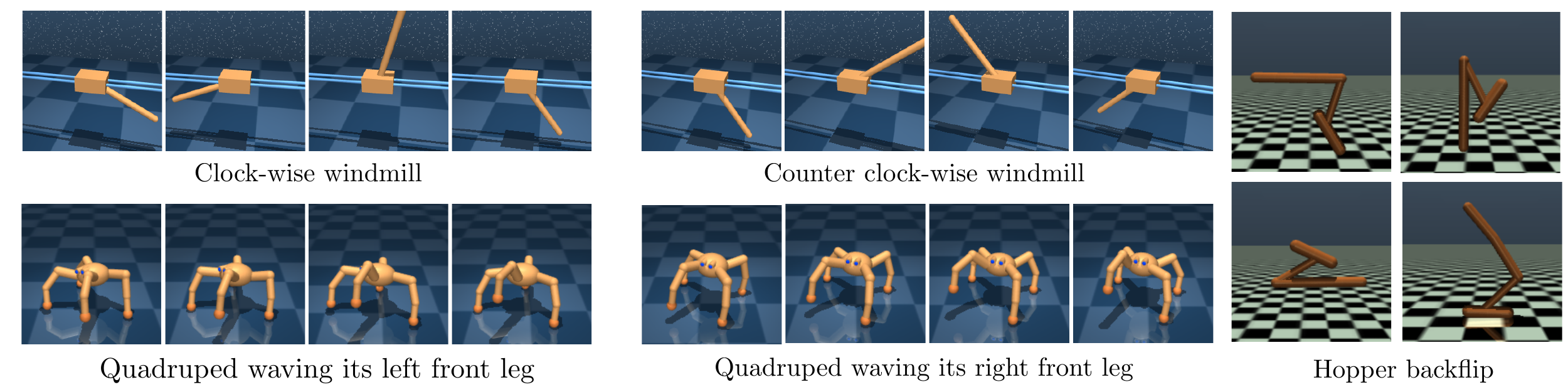}
\caption{Novel behaviors trained using feedback from human trainers. The corresponding videos and examples of selected queries are available at the supplementary material.}
\label{fig:novel_behavior}
\end{figure*}

\begin{figure} [t!] \centering
\subfigure[Agent trained with human preference]
{
\includegraphics[width=0.45\textwidth]{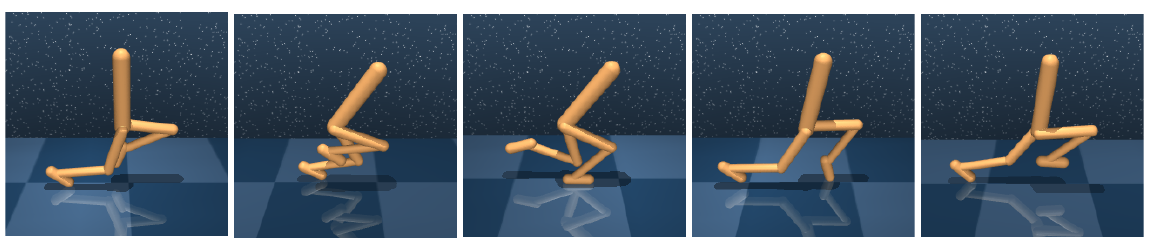}
\label{fig:reward_exploitation_walker_human}} 
\subfigure[Agent trained with hand-engineered reward]
{
\includegraphics[width=0.45\textwidth]{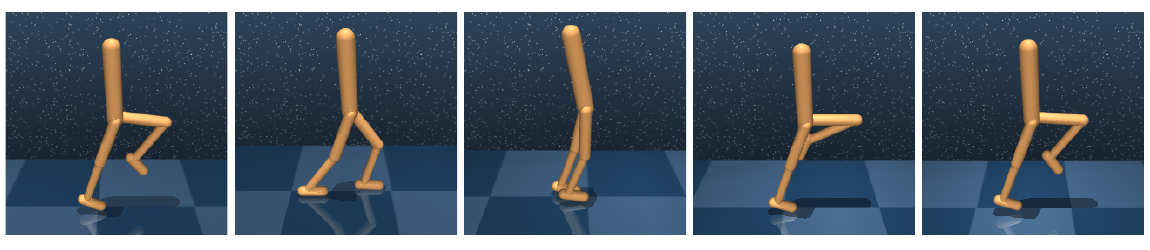}
\label{fig:reward_exploitation_walker_taskreward}}
\caption{Five frames from agents trained with (a) human preference and (b) hand-engineered reward from DMControl benchmark.}
\vspace{-0.1in}
\label{fig:reward_exploitation_walker}
\end{figure}

For evaluation, we compare to~\citet{preference_drl}, which is the current state-of-the-art approach using the same type of feedback. The primary differences in our method are (1) the introduction of unsupervised pre-training, (2) the accommodation of \emph{off-policy} RL, and (3) entropy-based sampling. 
We re-implemented~\citet{preference_drl} using the state-of-the-art on-policy RL algorithm: PPO~\citep{ppo}. 
We use the same reward learning framework and ensemble disagreement-based sampling as they proposed. 
We refer to this baseline as Preference PPO.

As an upper bound, 
since we evaluate against the task reward function, we also compare to SAC~\citep{sac} and PPO using the same ground truth reward.
For our method,
we pre-train an agent for 10K timesteps
and include these pre-training steps in all learning curves.
We do not alter any hyperparameters of the original SAC algorithm and use an ensemble of three reward models.
Unless stated otherwise, we use entropy-based sampling.
More experimental details including model architectures, sampling schemes, and reward learning are in the supplementary material.


\subsection{Benchmark Tasks with Unobserved Rewards}

{\bf Locomotion tasks from DMControl}.
Figure~\ref{fig:main_locomotion} shows the learning curves of \metabbr~with 1400, 700 or 400 pieces of feedback\footnote{One piece of feedback corresponds to one preference query.} and that of Preference PPO with 2100 or 1400 pieces of feedback on three complex environments: Cheetah-run, Walker-walk and Quadruped-walk.
Note that we explicitly give preference PPO an advantage by providing it with more feedback.
We find that given a budget of 1400 queries, \metabbr~(green) reaches the same performance as SAC (pink) while Preference PPO (purple) is unable to match PPO (black).
That PEBBLE requires less feedback than Preference PPO to match its respective oracle performance corroborates that PEBBLE is indeed more feedback-efficient.
These results demonstrate that \metabbr~can enable the agent to solve the tasks without directly observing the ground truth reward function.

For further analysis, we incorporated our pre-training with Preference PPO (red) and find that it improves performance for Quadruped and Walker. We emphasize that our insight of using pre-training is able to improve both methods in terms of feedback-efficiency and asymptotic performance, but PEBBLE is uniquely positioned to benefit as it is able to utilize unsupervised experience for policy learning.

{\bf Robotic manipulation tasks from Meta-world}. 
One application area in which HiL methods could have significant real-world impact is robotic manipulation, since learning often requires extensive engineering in the real world~\cite{dagps,pouring,em-rl, progressive-nets, akkaya2019solving, peng2020learning}. However, the common approach is to perform goal-conditioned learning with classifiers~\citep{singh2019end}, which can only capture limited information about what goal states \textit{are}, and not about \textit{how} they can be achieved. 
To study how we can utilize preference-based learning to perform more complex skills, we also consider six tasks covering a range of fundamental robotic manipulation skills from Meta-world (see Figure~\ref{fig:env_examples}).
As shown in Figure~\ref{fig:main_manipulation}, \metabbr~matches the performance of SAC using the ground truth reward and outperforms Preference PPO, given comparable (and more) feedback, on every task. By demonstrating the applicability of \metabbr~to learning a variety of robotic manipulation tasks, we believe that we are taking an important step towards anyone (non-experts included) being able to teach robots in real-world settings.

\subsection{Ablation Study}

{\bf Contribution of each technique}. 
In order to evaluate the individual effects of each technique in \metabbr, we incrementally apply unsupervised pre-training and relabeling. Figure~\ref{fig:ablation_unsuper} shows the learning curves of \metabbr~with 1400 queries on Quadruped-walk.
First, we remark that relabeling significantly improves performance because it enables the agent to be robust to changes in its reward model.
By additionally utilizing unsupervised pre-training, both sample-efficiency and asymptotic performance of \metabbr~are further improved because showing diverse behaviors to a teacher can induce a better-shaped reward. 
This shows that \metabbr's key ingredients are fruitfully wed, and their unique combination is crucial to our method's success.

{\bf Effects of sampling schemes}.
We also analyze the effects of different sampling schemes to select queries.
Figure~\ref{fig:ablation_sampling} shows the learning curves of \metabbr~with three different sampling schemes: uniform sampling, disagreement sampling and entropy sampling on Quadruped-walk. 
For this complex domain, we find that the uncertainty-based sampling schemes (using ensemble disagreement or entropy) are superior to the naive uniform sampling scheme.
However, we note that they did not lead to extra gains on relatively simple environments, like Walker and Cheetah, similar to observations from \citet{ibarz2018preference_demo} (see the supplementary material for more results). 

{\bf Comparison with step-wise feedback}.
We also measure the performance of \metabbr~by varying the length of segments.
Figure~\ref{fig:ablation_seg} shows that feedback from longer segments (green curve) provide more meaningful signal than step-wise feedback (red curve). We believe that this is because longer segments can provide more context in reward learning. 

\subsection{Human Experiments}

{\bf Novel behaviors}.
We show that agents can perform various novel behaviors based on human feedback using \metabbr~in Figure~\ref{fig:novel_behavior}. 
Specifically, we demonstrate (a) the Cart agent swinging a pole (using 50 queries), 
(b) the Quadruped agent waving a front leg (using 200 queries), 
and (c) the Hopper performing a backflip (using 50 queries).
We note that the human is indeed able to guide the agent in a controlled way, as evidenced by training the same agent to perform several variations of the same task (e.g., waving different legs or spinning in opposite directions).
The videos of all behaviors and examples of selected queries are available in the supplementary material.

{\bf Reward exploitation}.
One concern in utilizing hand-engineered rewards is that an agent can exploit unexpected sources of reward, leading to unintended behaviors.
Indeed, we find that the Walker agent learns to walk using only one leg even though it achieves the maximum scores as shown in Figure~\ref{fig:reward_exploitation_walker_taskreward}.
However, using 200 human queries, we were able to train the Walker to walk in a more natural, human-like manner (using both legs) as shown in Figure~\ref{fig:reward_exploitation_walker_human}.
This result clearly shows the advantage of HiL RL to avoid reward exploitation.


\section{Discussion}

In this work, we present \metabbr, a feedback-efficient algorithm for HiL RL.
By leveraging unsupervised pre-training and off-policy learning, 
we show that sample- and feedback-efficiency of HiL RL can be significantly improved and this framework can be applied to tasks of higher complexity than previously considered by previous methods, including a variety of locomotion and robotic manipulation skills.
Additionally, we demonstrate that \metabbr~can learn novel behaviors and avoid reward exploitation, leading to more desirable behaviors compared to an agent trained with respect to an engineered reward function.
We believe by making preference-based learning more tractable, \metabbr~may facilitate broadening the impact of RL beyond settings in which experts can carefully craft reward functions to those in which laypeople can likewise utilize the advances of robot learning in the real world.

\section*{Acknowledgements}
This research is supported in part by 
ONR PECASE N000141612723, 
NSF NRI \#2024675,
Tencent, 
and Berkeley Deep Drive.
Laura Smith was supported by NSF Graduate Research Fellowship.
We thank Abhishek Gupta, Joey Hejna, Qiyang (Colin) Li, Fangchen Liu, Olivia Watkins, and Mandi Zhao for providing helpful feedbacks and suggestions.
We also thank anonymous reviewers for critically reading the manuscript and suggesting substantial improvements.

\bibliography{reference}
\bibliographystyle{icml2021}

\clearpage
\appendix
\onecolumn

\begin{center}{\bf {\LARGE Appendix}}
\end{center}

\section{State Entropy Estimator} \label{app:state_entropy}

To approximate state entropy, we employ the simplified version of particle-based entropy estimator~\cite{beirlant1997nonparametric, singh2003nearest}.
Specifically, let $\state$ be a random variable with a probability density function $p$ whose support is a set $\mathcal{S} \subset \mathbb{R}^{q}$. 
Then its differential entropy is given as $\mathcal{H}(\state) = -\mathbb{E}_{\state \sim p(\state)}[\log p(\state)]$.
When the distribution $p$ is not available, this quantity can be estimated given $N$ i.i.d realizations of $\{\state_{i}\}^{N}_{i=1}$ \cite{beirlant1997nonparametric}.
However, since it is difficult to estimate $p$ with high-dimensional data, particle-based $k$-nearest neighbors ($k$-NN) entropy estimator \cite{singh2003nearest} can be employed:
\begin{align}
    \widehat{\mathcal{H}}(\state) &= \frac{1}{N} \sum^{N}_{i=1} \log \frac{N \cdot ||\state_{i} - \state_{i}^{k}||_{2}^{q} \cdot {\widehat \pi}^{\frac{q}{2}}}{k \cdot \Gamma(\frac{q}{2} + 1)} + C_{k}
    \label{eq:knn_state_entropy_estimator}\\
    &\propto \frac{1}{N}\sum^{N}_{i=1} \log ||\state_{i} - \state_{i}^{k}||_{2}, 
    \label{eq:simplied_knn_state_entropy_estimator}
 \end{align}
where $\widehat \pi$ is the ratio of a circle's circumference to its diameter, $\state_{i}^{k}$ is the $k$-NN of $\state_{i}$ within a set $\{\state_{i}\}_{i=1}^{N}$, $C_{k} = \log k - \Psi(k)$ a bias correction term, $\Psi$ the digamma function, $\Gamma$ the gamma function, $q$ the dimension of $\state$, and the transition from (\ref{eq:knn_state_entropy_estimator}) to (\ref{eq:simplied_knn_state_entropy_estimator}) always holds for $q > 0$. 
Then, from \autoref{eq:simplied_knn_state_entropy_estimator},
we define the intrinsic reward of the current state $\state_t$ as follows:
\begin{align*}
    r^{\text{\tt int}}(\state_t) = \log (|| \state_{t} - \state_{t}^{k}||).
\end{align*}

\begin{table}[h!]
\begin{center}
\resizebox{\columnwidth}{!}{
\begin{tabular}{ll|ll}
\toprule
\textbf{Hyperparameter} & \textbf{Value} &
\textbf{Hyperparameter} & \textbf{Value} \\
\midrule
Initial temperature & $0.1$ & Hidden units per each layer & 1024 (DMControl), 256 (Meta-world) \\ 
Learning rate  & $0.0003$ (Meta-world), $0.001$ (cheetah) &  Batch Size  & $1024$ (DMControl), $512$ (Meta-world)\\ 
& $0.0001$ (qauadruped), $0.0005$ (walker) &  Optimizer  & Adam~\citep{kingma2014adam}\\ 
Critic target update freq & $2$  & Critic EMA $\tau$ & $0.005$ \\ 
$(\beta_1,\beta_2)$  & $(.9,.999)$ & Discount $\gamma$ & $.99$\\
\bottomrule
\end{tabular}}
\end{center}
\caption{Hyperparameters of the SAC algorithm. Most hyperparameters values are unchanged across environments with the exception for learning rate.}
\label{table:hyperparameters_sac}
\end{table}

\begin{table}[h!]
\begin{center}
\resizebox{\columnwidth}{!}{
\begin{tabular}{ll|ll}
\toprule
\textbf{Hyperparameter} & \textbf{Value} &
\textbf{Hyperparameter} & \textbf{Value} \\
\midrule
GAE parameter $\lambda$ & 0.92 & Hidden units per each layer & 1024 (DMControl), 256 (Meta-world) \\ 
Learning rate  & $0.0003$ (Meta-world), $0.0001$ (quadruped) &  Batch Size  & $512$ (cheetah), $128$ (Otherwise)\\ 
&  $5e^{-5}$ (quadruped, Walker) &  \# of timesteprs per rollout & 100 (cheetah, Walker), 500 (quadruped) \\ 
\# of environments per worker & 16 (quadruped, cheetah), 32 (Walker)  & PPO clip range & 0.2 \\ 
Entropy bonus & 0.0 & Discount $\gamma$ & $.99$\\
\bottomrule
\end{tabular}}
\end{center}
\caption{Hyperparameters of the PPO algorithm. Most hyperparameters values are unchanged across environments with the exception for learning rate.}
\label{table:hyperparameters_ppo}
\end{table}

\section{Experimental Details} \label{app:exp_setup}

{\bf Training  details}. For our method, we use the publicly released implementation repository of the SAC algorithm (\url{https://github.com/denisyarats/pytorch_sac}) with a full list of hyperparameters in Table~\ref{table:hyperparameters_sac}. On the DMControl environments, we use segments of length 50 and a frequency of teacher feedback ($K$ in Algorithm~\ref{alg:pebble}) of 20K timesteps, which corresponds to roughly 20 episodes. 
We choose the number of queries per feedback session $M=140, 70, 40$ for the maximum budget of 1400, 700, 400 on Walker and Cheetah, and choose $M=70, 35, 20$ for the maximum budget of 1400, 700, 400 on Quadruped.
For Meta-world, we use segments of length 10 and set $M=64, K=2400$ for the maximum budget of 2500, 5000, and 10000 on Drawer Close, Window Open, Door Open, and Button Press and $M=128, K=4800$ for maximum budget of 25000, 50000 on Sweep Into and Drawer Open.

For preference PPO, we use the publicly released implementation repository of the PPO algorithm (\url{https://github.com/DLR-RM/stable-baselines3}) with a full list of hyperparameters in Table~\ref{table:hyperparameters_ppo}.
We choose the number of queries per feedback session $M=70, 45$ for the maximum budget of 2100, 1400 on the DMControl environments.
For the reward model, we use same setups for our method. For Meta-world, we use segments of length 10 and set $M=256, K=2400$ for all environments and budgets of feedback.

{\bf Reward model}. For the reward model, 
we use a three-layer neural network with 256 hidden units each, using leaky ReLUs. 
To improve the stability in reward learning, 
we use an ensemble of three reward models, and bound the output using tanh function.
Each model is trained by optimizing the cross-entropy loss defined in \eqref{eq:reward-bce} using ADAM learning rule~\citep{kingma2014adam} with the initial learning rate of 0.0003.

{\bf Environments}. 
We follow the standard evaluation protocol for the benchmark locomotion tasks from DMControl. 
The Meta-world single-task benchmark involves training and testing on a single instantiation (fixed reset and goal) of the task. To constitute a more realistic single-task manipulation setting, 
we randomize the reset and goal positions in all our experiments. 
We also use new reward function, which are nicely normalized and make the tasks stable.


\section{Effects of Sampling Schemes} \label{app:exp_sampling}

\begin{figure*} [h!] \centering
\subfigure[Quadruped]
{
\includegraphics[width=0.31\textwidth]{media/arxiv_sampling_quadruped_walk.pdf}
\label{fig:app_sampling_quad}}
\subfigure[Walker]
{
\includegraphics[width=0.31\textwidth]{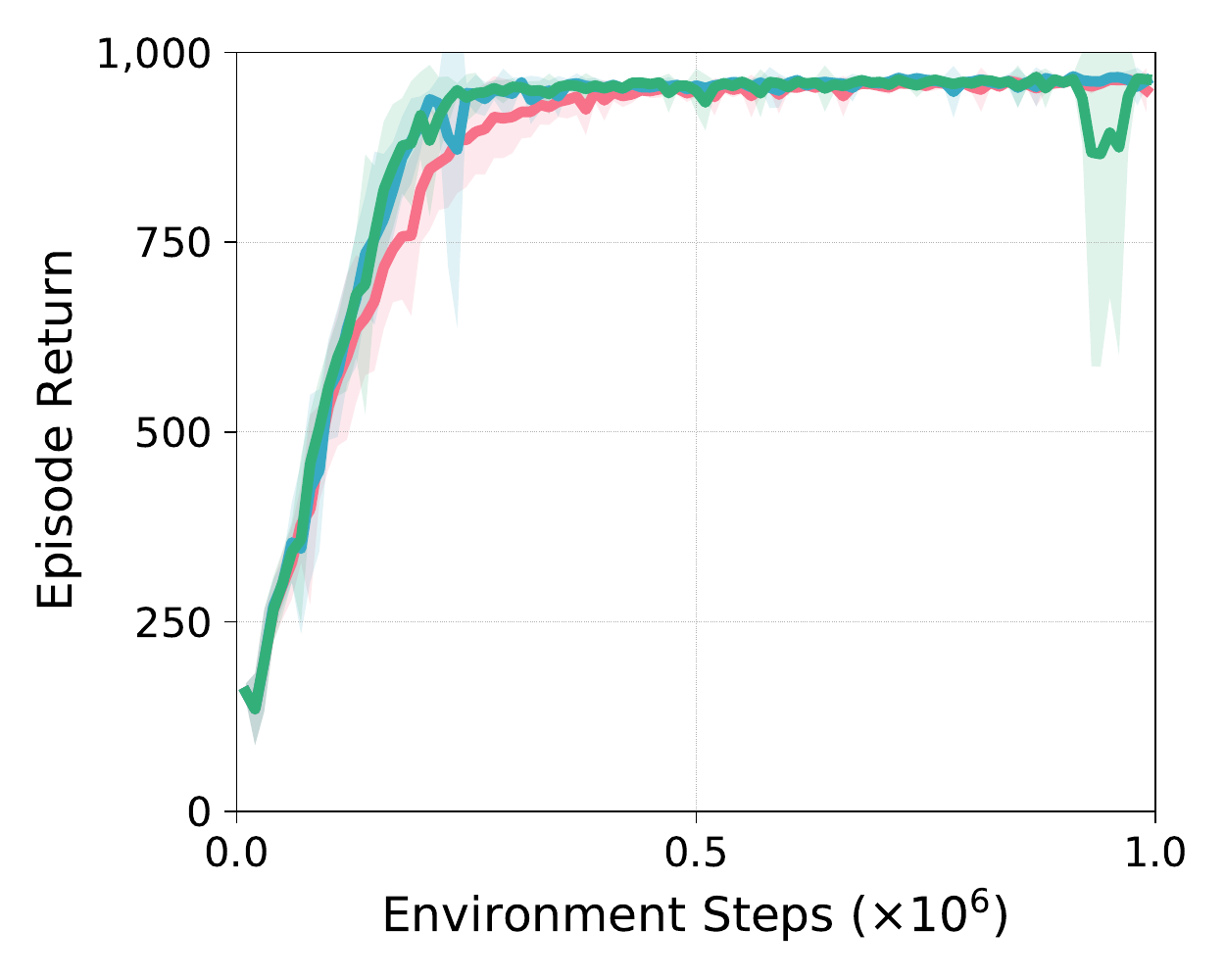}
\label{fig:app_sampling_walker}} 
\subfigure[Cheetah]
{
\includegraphics[width=0.31\textwidth]{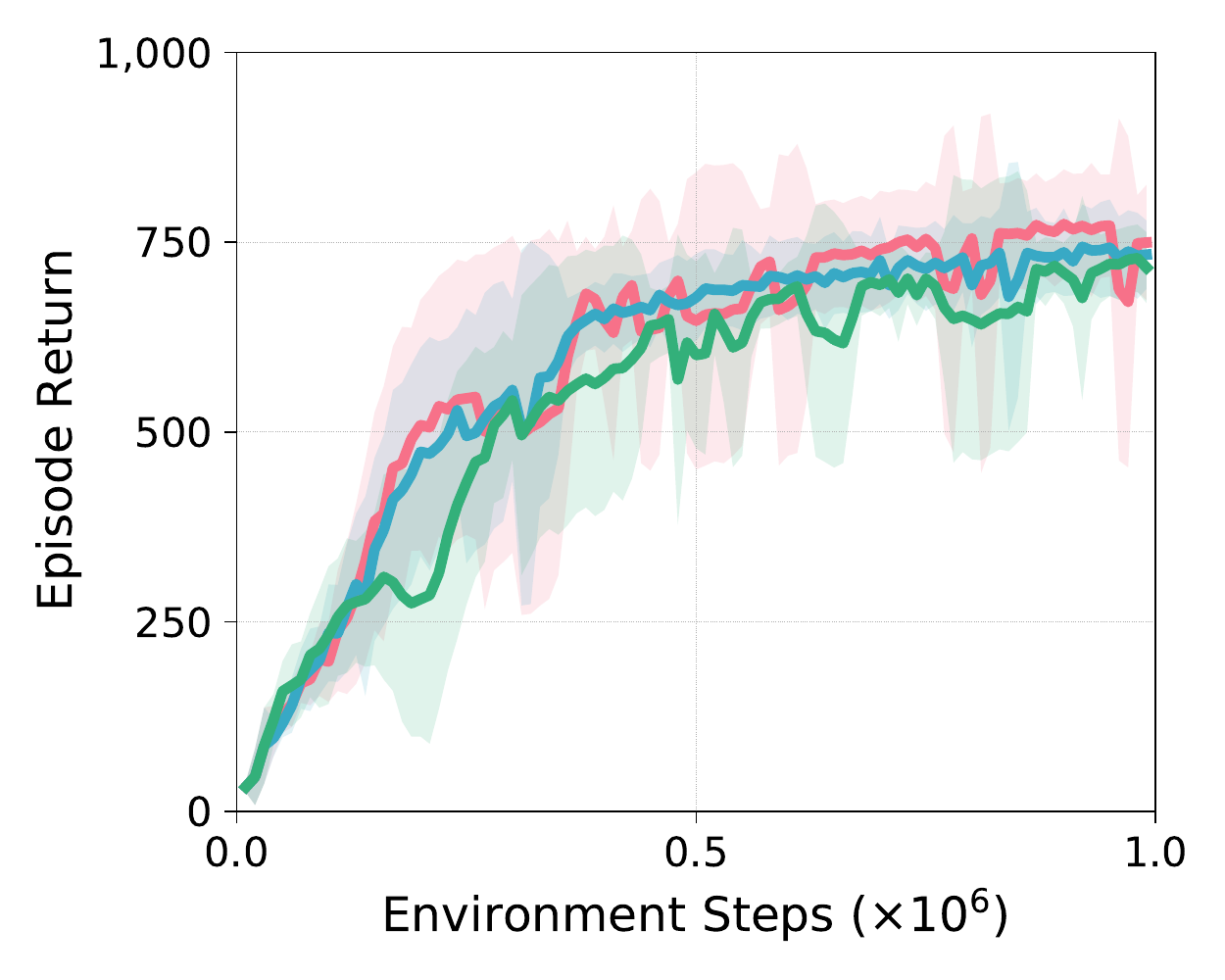}
\label{fig:app_sampling_cheetah}}
\caption{Learning curves of \metabbr~with 1400 pieces of feedback by varying sampling schemes. 
The solid line and shaded regions represent the mean and standard deviation, respectively, across ten runs.}
\label{fig:app_sampling}
\end{figure*}

\begin{figure*}[h!]
    \centering
    \includegraphics[width=0.325\linewidth]{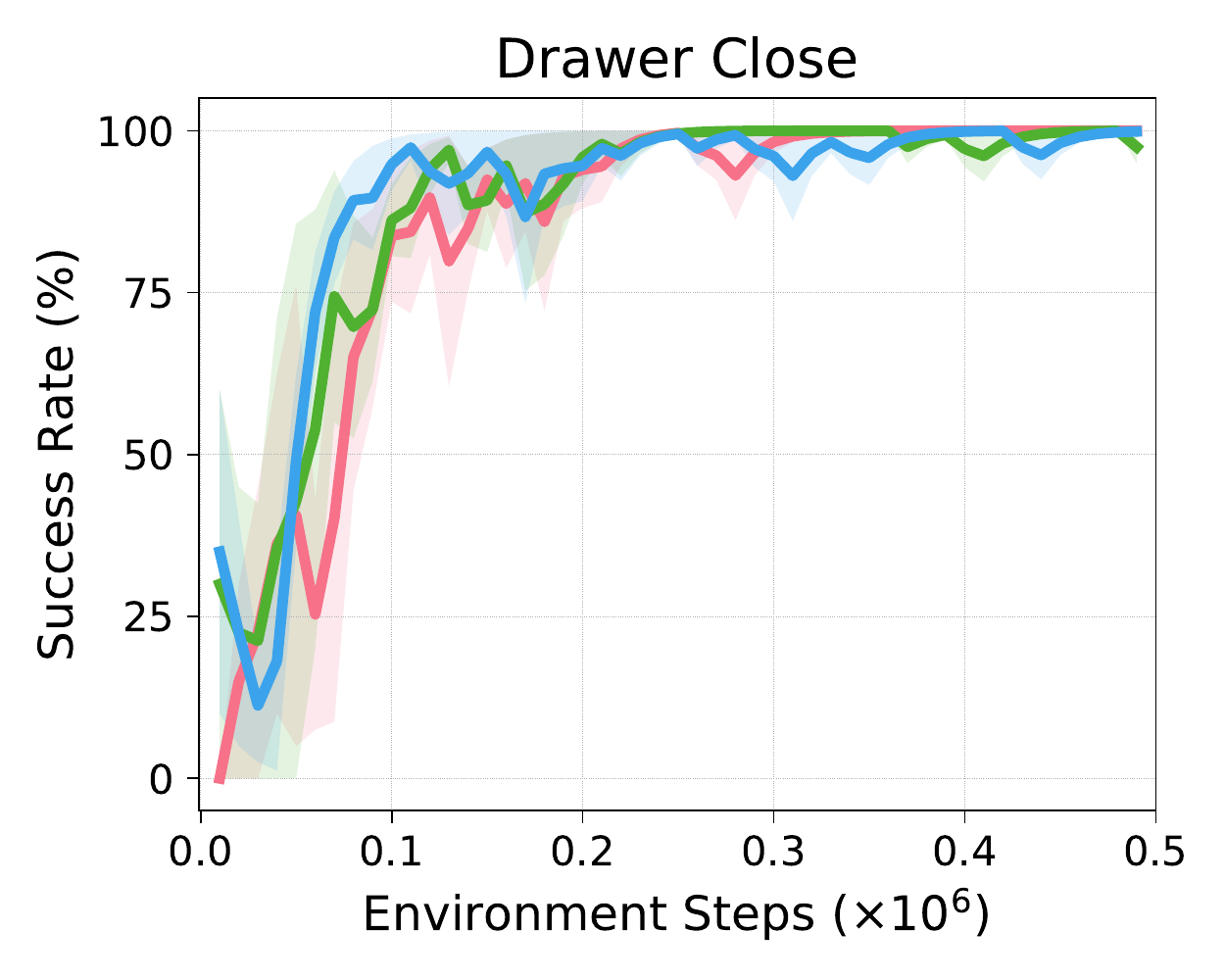} 
    \includegraphics[width=0.325\linewidth]{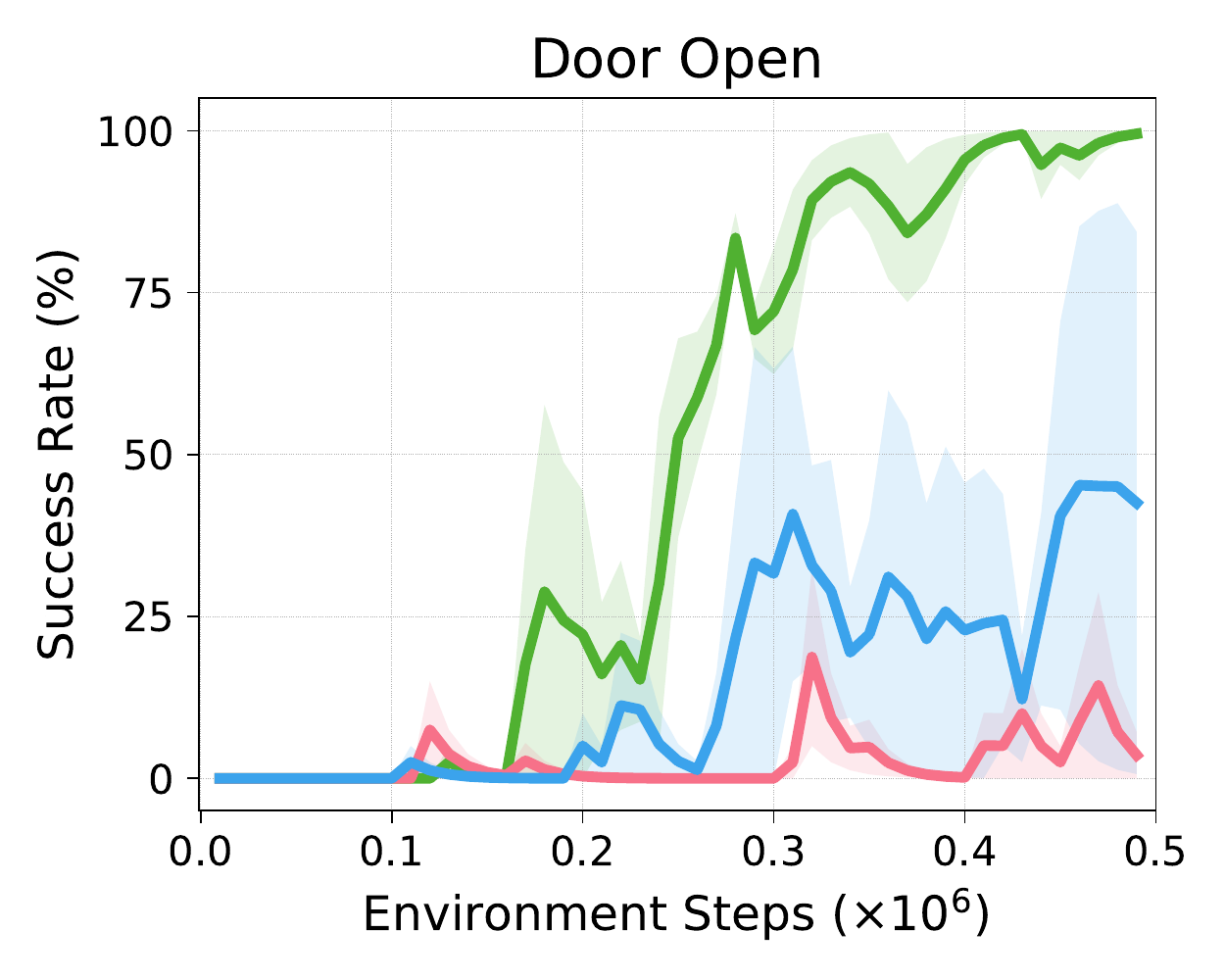} 
    \includegraphics[width=0.325\linewidth]{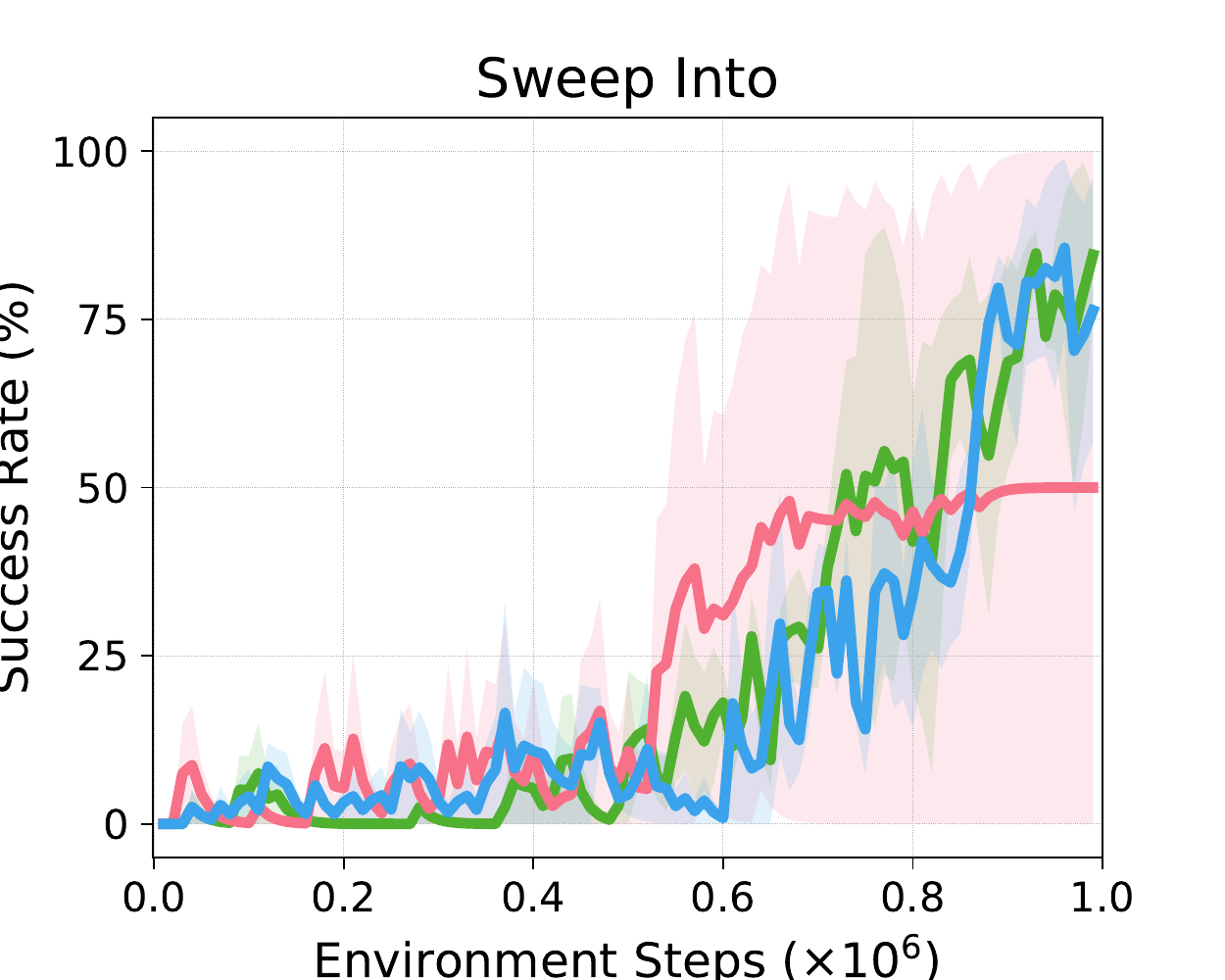} \\
    \includegraphics[width=0.325\linewidth]{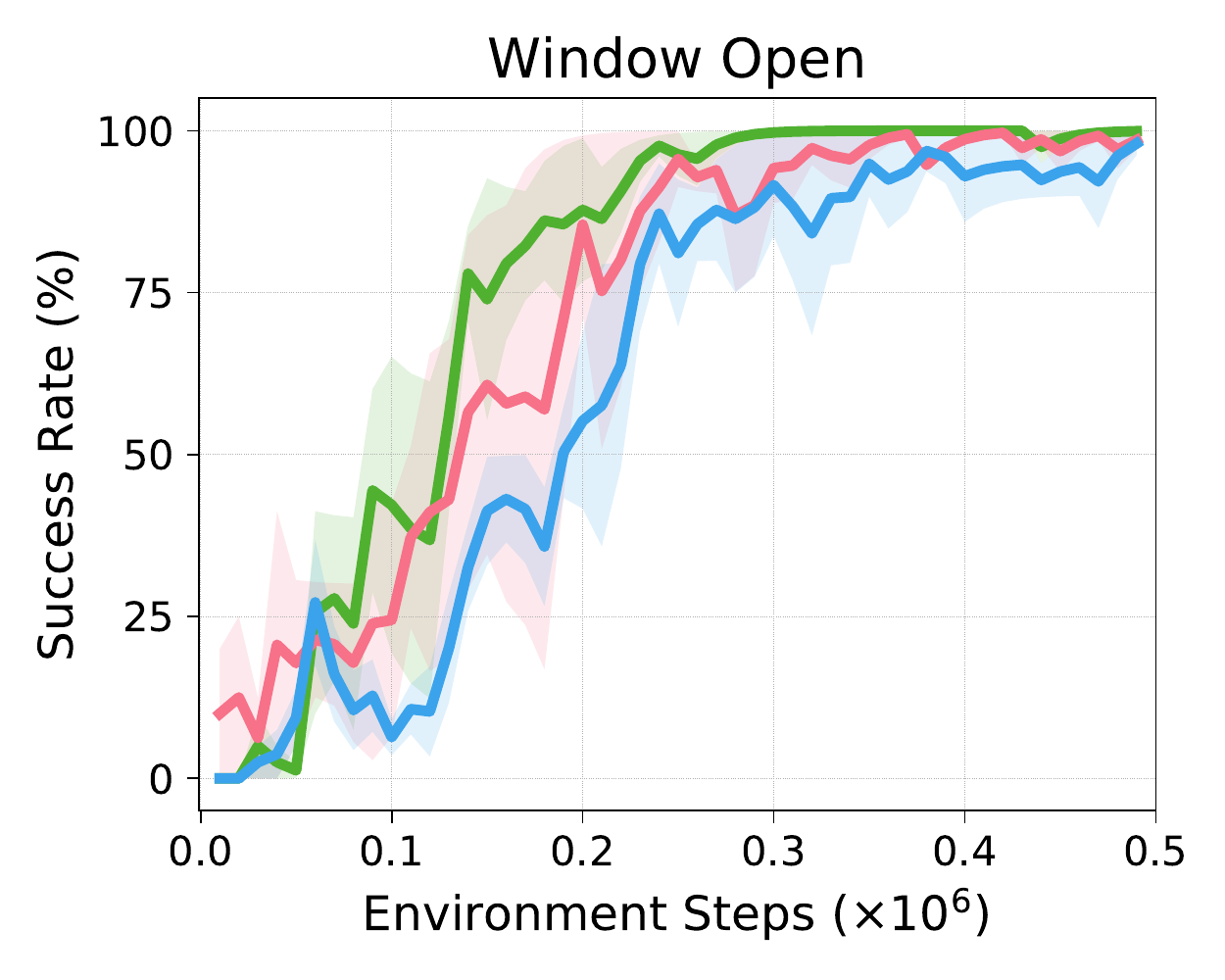}
    \includegraphics[width=0.325\linewidth]{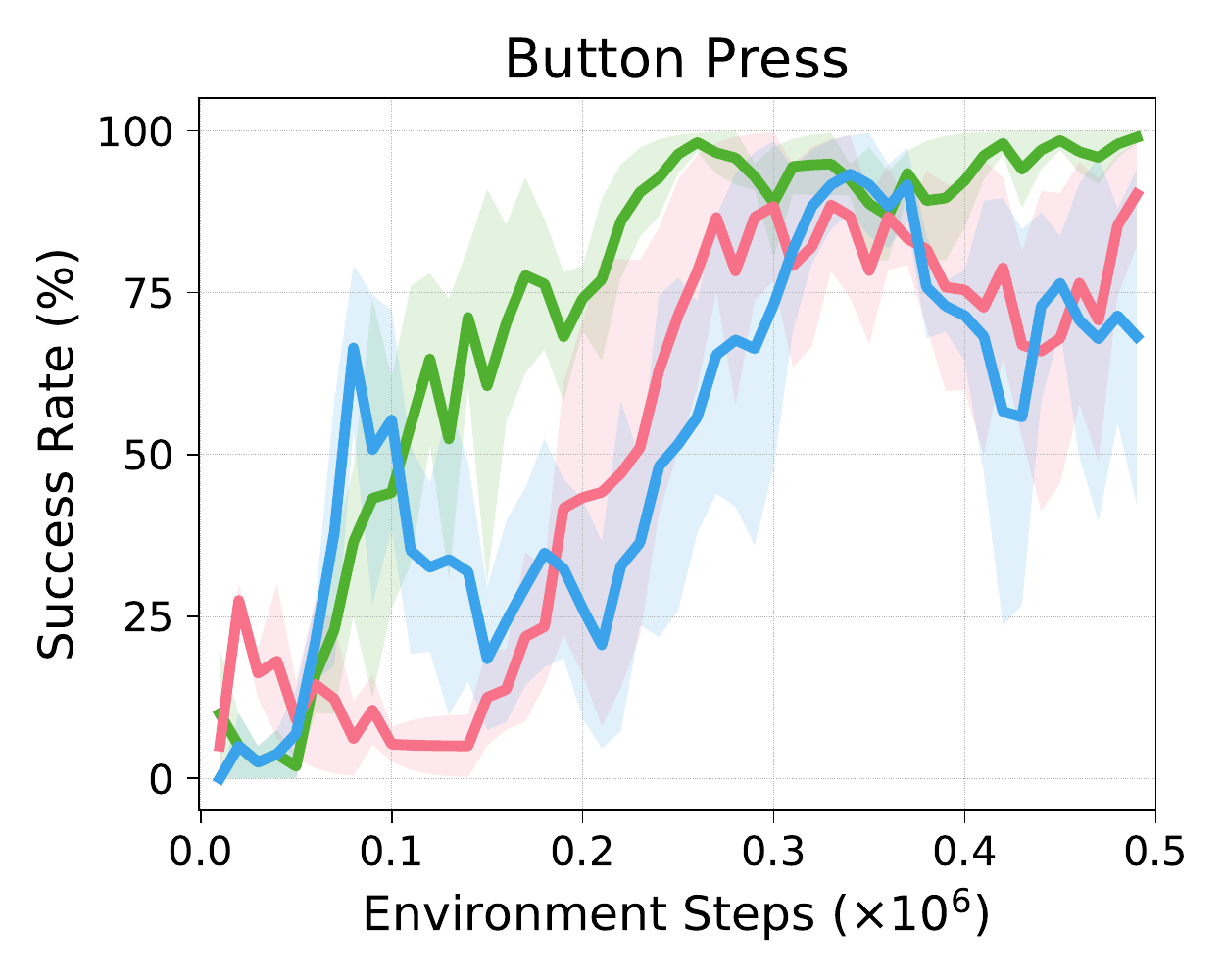}
    \includegraphics[width=0.325\linewidth]{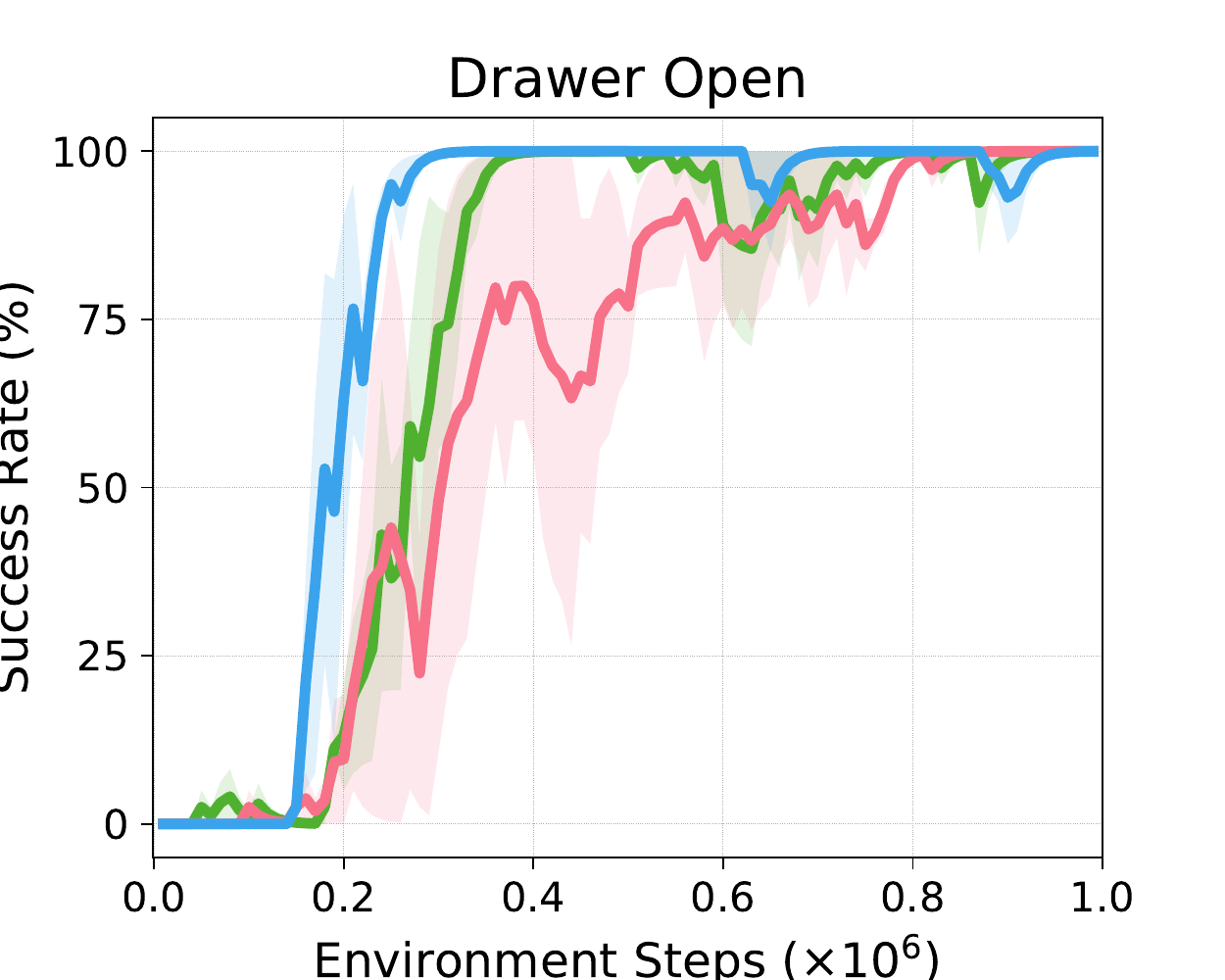}\\
    \vspace{.1cm}
    \includegraphics[width=0.7\linewidth]{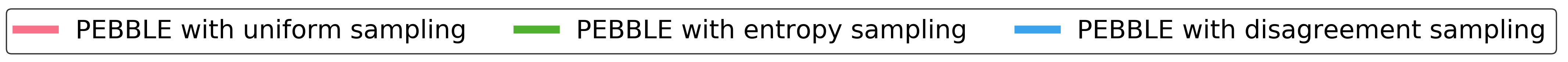} 
    \vspace{-.2cm}
    \caption{Learning curves of \metabbr~with various sampling schemes on the Meta-world tasks. The solid line and shaded regions represent the mean and standard deviation, respectively, across ten runs.}
    \label{fig:app_sampling_manipulation}
    \vspace{-.5cm}
\end{figure*}

Figures~\ref{fig:app_sampling} and \ref{fig:app_sampling_manipulation} show the learning curves of \metabbr~with various sampling schemes. For Quadruped, we find that the uncertainty-based sampling schemes (using ensemble disagreement or entropy) are superior to the naive uniform sampling scheme. However, they did not lead to extra gains on relatively simple environments, like Walker and Cheetah, similar to observations from \citet{ibarz2018preference_demo}. 
Similarly, on the robotic manipulation tasks, 
we find little difference in performance for simpler tasks (Drawer Close, Window Open). 
However, we find that the uncertainty-based sampling schemes generally fare better on the other environments.

\section{Examples of Selected Queries}

\begin{figure*} [h!] \centering
\subfigure[Clock-wise windmill]
{
\includegraphics[width=0.47\textwidth]{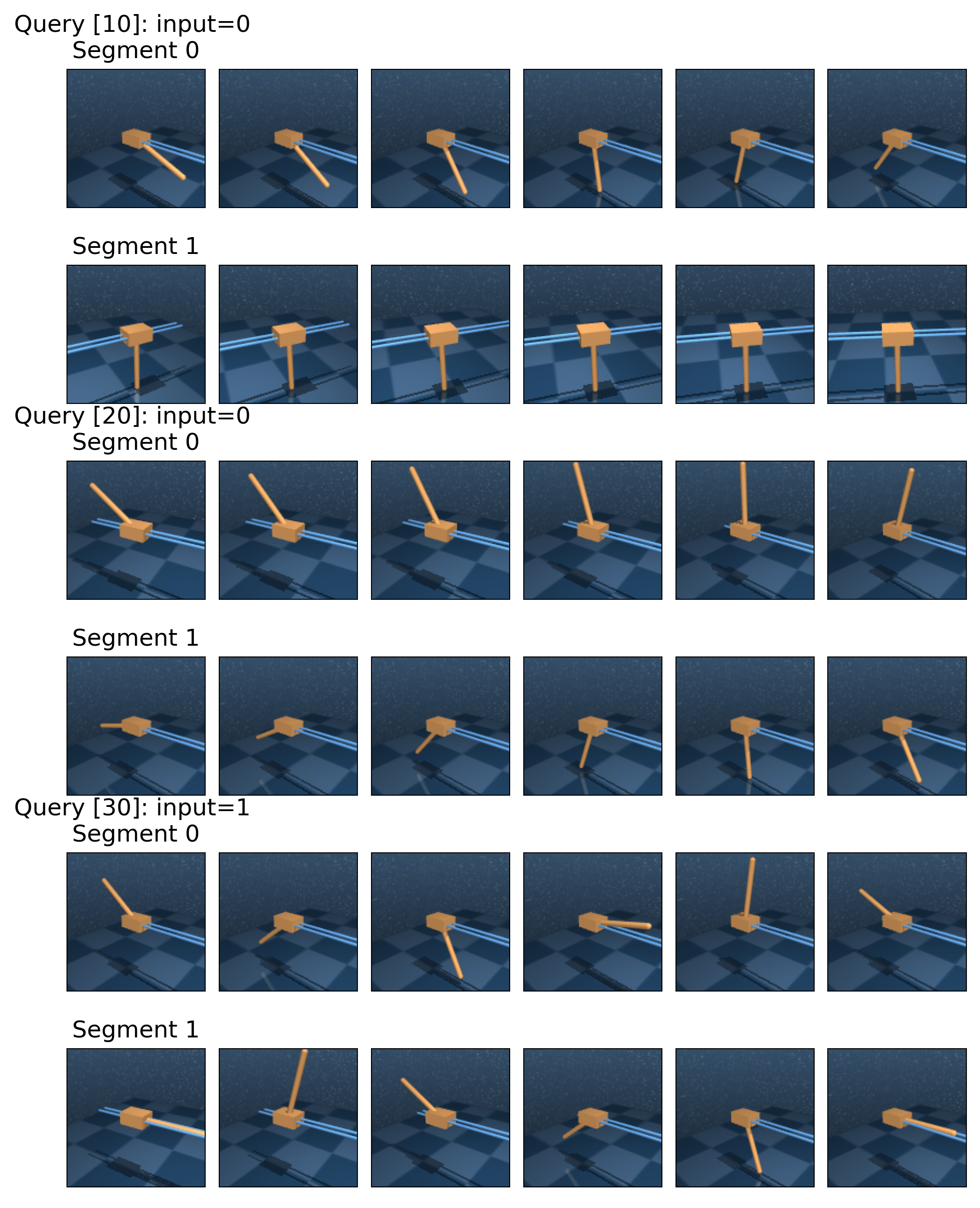}
\label{fig:query_cart_clock}} 
\subfigure[Counter clock-wise windmill]
{\includegraphics[width=0.47\textwidth]{media/query_cart_clock_wise.png}
\label{fig:query_cart_counter_clock}}
\caption{Examples from the selected queries to teach the Cart agent.}
\label{fig:query_cart}
\end{figure*}

\begin{figure*} [h!] \centering
\subfigure[Waving left front leg]
{
\includegraphics[width=0.47\textwidth]{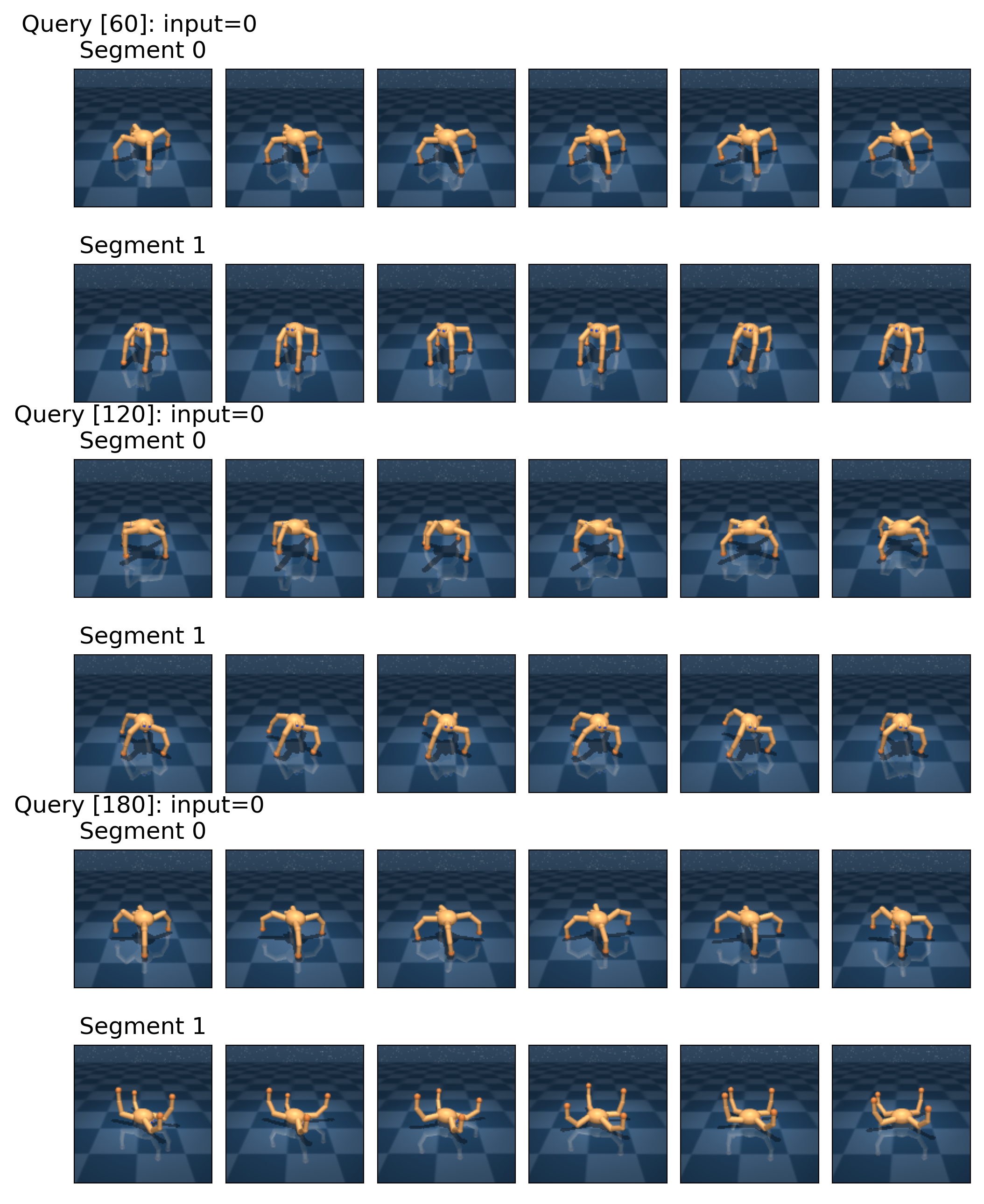}
\label{fig:query_qudruped_left}} 
\subfigure[Waving right front leg]
{\includegraphics[width=0.47\textwidth]{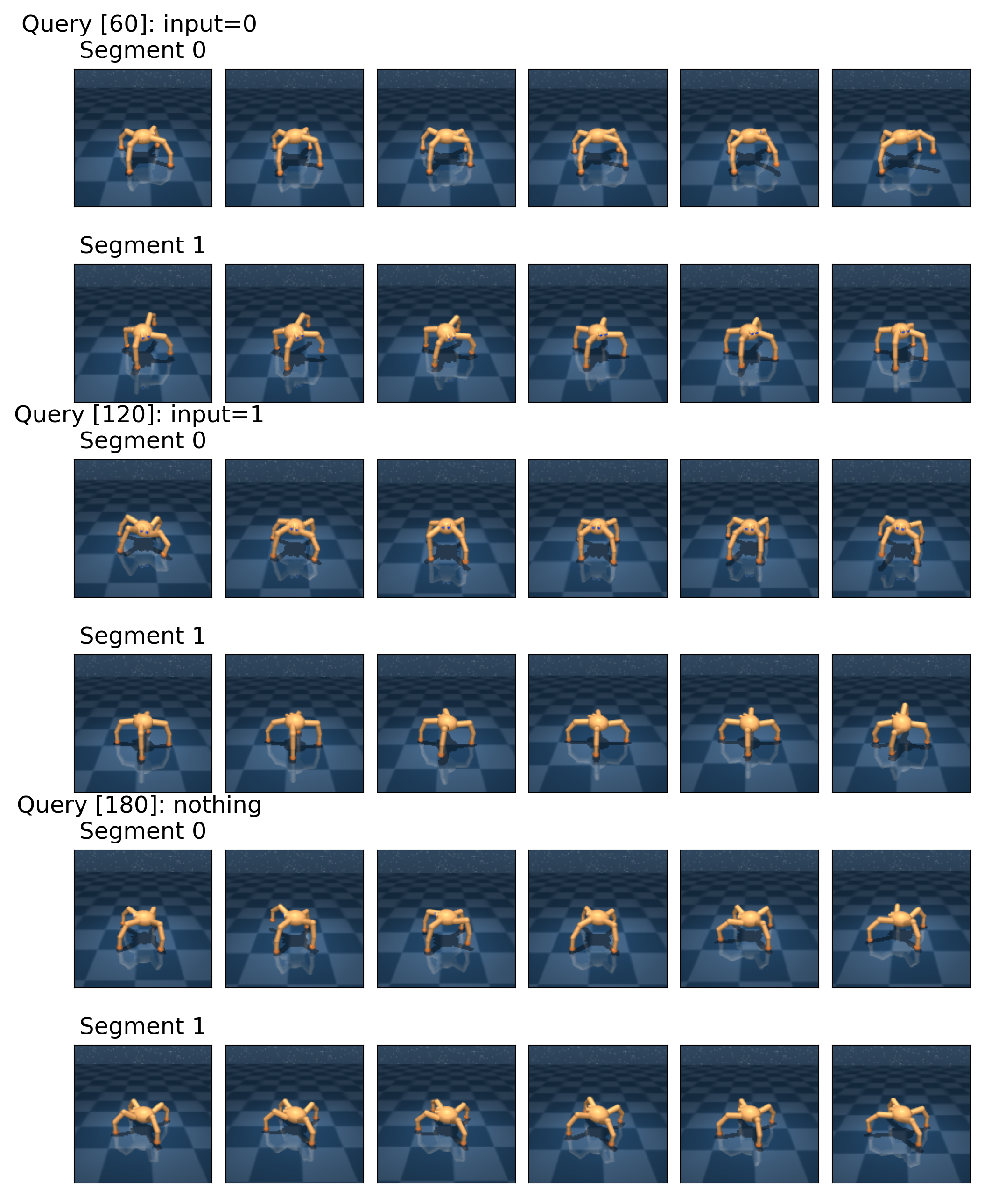}
\label{fig:query_qudruped_right}}
\caption{Examples from the selected queries to teach the Quadruped agent.}
\label{fig:query_qudruped}
\end{figure*}

\begin{figure*}
\centering
\begin{tabular}{cc}
\includegraphics[width=0.46\linewidth]{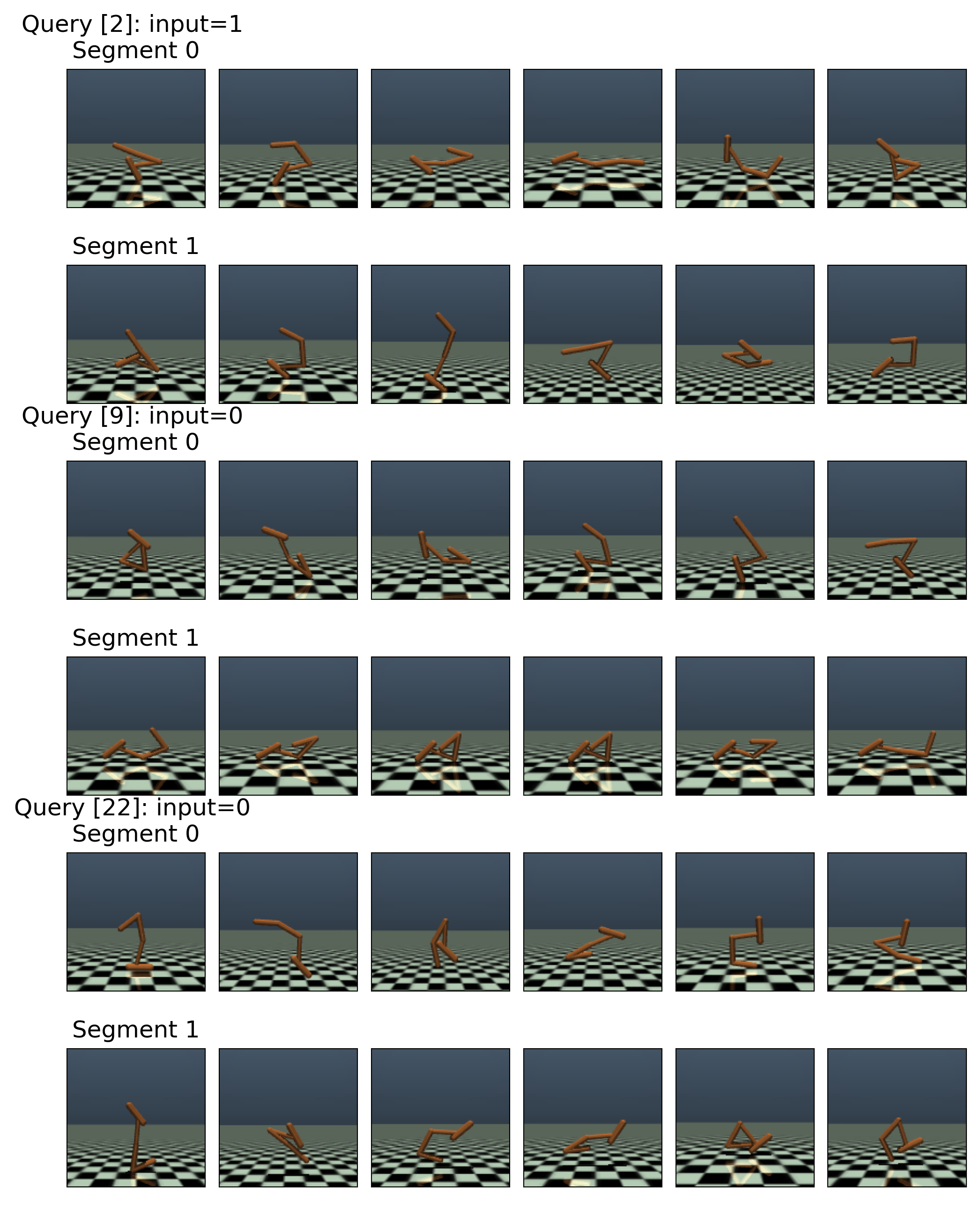}
& \includegraphics[width=0.46\linewidth]{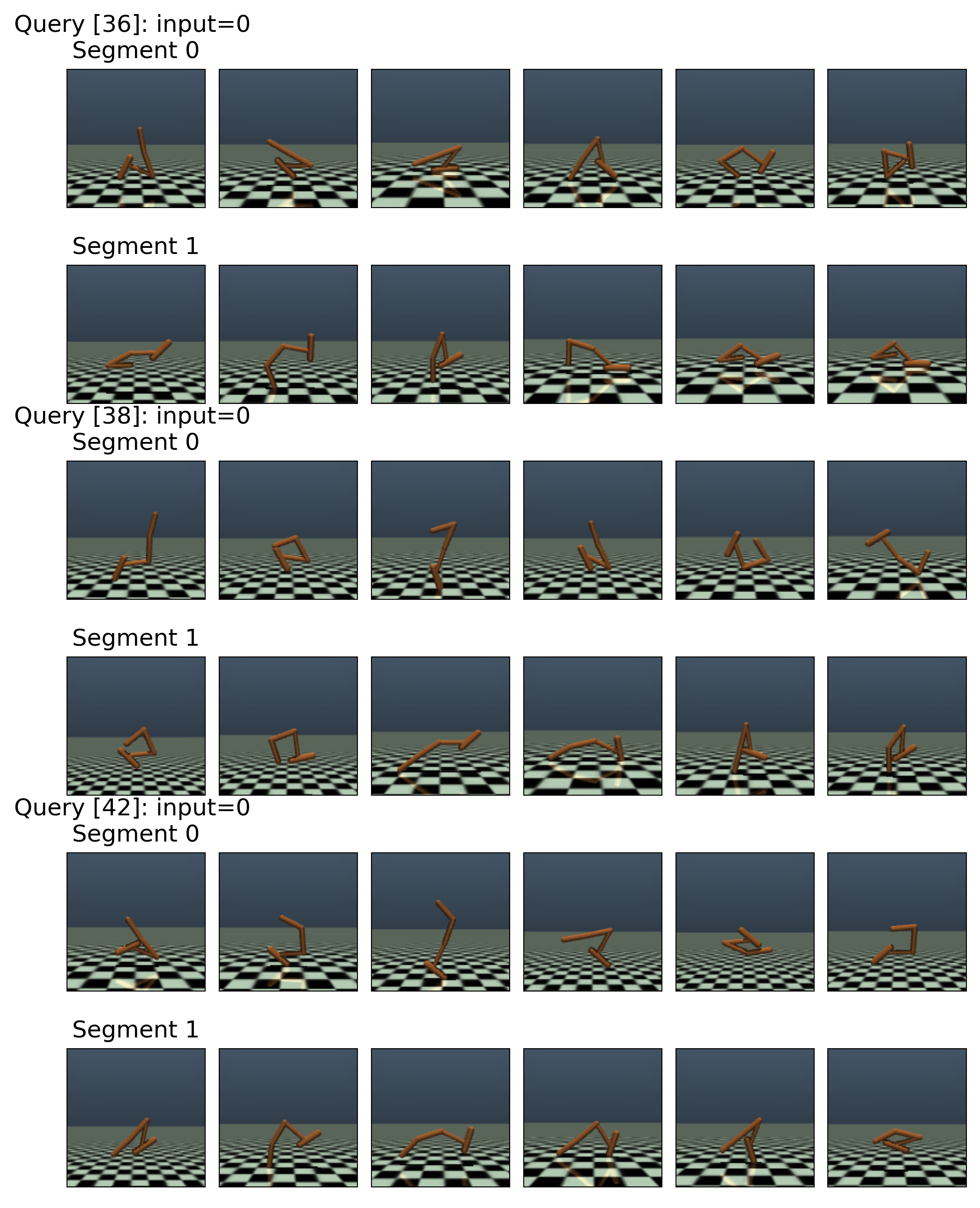}
\end{tabular}
\caption{Examples from the selected queries to teach the Hopper agent.}
\label{fig:query_hopper}
\end{figure*}

Figure~\ref{fig:query_cart}, \ref{fig:query_qudruped} and \ref{fig:query_hopper} show some examples from the selected queries to teach the agents. 


\end{document}